\documentclass[journal]{IEEEtran}

\usepackage{graphicx}
\usepackage{amsfonts,amssymb}
\usepackage{lipsum}
\usepackage{bm}
\usepackage{epstopdf}
\usepackage{subeqnarray}
\usepackage{algorithm}
\usepackage{float}
\usepackage{algorithmic}
\usepackage{subfigure}
\usepackage{tabularx}
\usepackage{amsmath}
\usepackage{booktabs }
\usepackage{amsthm}
\usepackage{hyperref}
\usepackage{url}
\usepackage{cite}
\usepackage{multirow}
\graphicspath{{figures/}}
\usepackage{array}
\newcommand{\PreserveBackslash}[1]{\let\temp=\\#1\let\\=\temp}
\newcolumntype{C}[1]{>{\PreserveBackslash\centering}p{#1}}
\newcolumntype{R}[1]{>{\PreserveBackslash\raggedleft}p{#1}}
\newcolumntype{L}[1]{>{\PreserveBackslash\raggedright}p{#1}}
\usepackage{caption2}

\ifCLASSINFOpdf

\else

\fi

\begin{document}

\title{Differentially Private Online Learning\\ for Cloud-Based Video Recommendation with Multimedia Big Data in Social Networks }

\author{\normalsize Pan Zhou\dag, \emph{Member, IEEE}, \normalsize Yingxue Zhou\dag, \emph{Student Member, IEEE},  Dapeng Wu, \emph{Fellow, IEEE} and Hai Jin, \emph{Senior Member, IEEE} \IEEEcompsocitemizethanks{\IEEEcompsocthanksitem Pan Zhou and Yingxue Zhou, \dag These authors contributed equally to this work and are considered co-first authors, are with the School
of Electronic Information and Communications, Huazhong University of
Science and Technology, Wuhan 430074, China.
E-mail: hustzhouyx@gmail.com, panzhou@hust.edu.cn
\IEEEcompsocthanksitem Dapeng Wu is with the Department of Electrical and Computer Engineering, University of Florida.
E-mail: wu@ece.ufl.edu
\IEEEcompsocthanksitem Hai Jin is with the School of Computer Science and Technology,
Huazhong University of Science and Technology,
Wuhan, 430074, China.  E-mail: hjin@hust.edu.cn}
\thanks{Manuscript received XXXXX; revised XXXXX.}
}

\markboth{IEEE TRANSACTIONS ON MULTIMEDIA,~Vol.~X, No.~X, September~20XX}%
{Shell \MakeLowercase{\textit{et al.}}: Bare Demo of IEEEtran.cls for Journals}

\maketitle

\begin{abstract}
With the rapid growth in multimedia services and the enormous offers of video contents in online social networks, users have difficulty in obtaining their interests. Therefore, various  personalized recommendation systems have been proposed. However, they ignore that the  accelerated proliferation of social media data has led to the big data era, which has greatly impeded the process of video recommendation. In addition, none of them has considered both the privacy of users' contexts (e,g., social status, ages and hobbies) and video service vendors' repositories, which are extremely sensitive and of significant commercial value. To handle the problems, we propose a cloud-assisted differentially private video recommendation system based on distributed online learning. In our framework, service vendors are modeled as distributed cooperative learners, recommending videos according to user's context, while simultaneously adapting the video-selection strategy based
on user-click feedback to maximize total user clicks (reward). Considering the sparsity and heterogeneity of big social media data, we also propose a novel \emph{geometric differentially private} model, which can greatly reduce the performance (recommendation accuracy) loss.  Our simulation shows the proposed algorithms outperform other existing methods and keep a delicate balance between computing accuracy and privacy preserving level.
\end{abstract}

\begin{IEEEkeywords}
Online social networks, multimedia big data, video recommendation, distributed online learning, differential privacy, media cloud.
\end{IEEEkeywords}

\IEEEpeerreviewmaketitle

\section{Introduction}

\IEEEPARstart{I}{n} recent years, online social networks (OSNs) have been massively growing, where users can share and consume all kinds of multimedia contents. As a result, given the numerous different genres of videos in social media, how to discover the videos of personal interest and recommend them to individual users are of great significance. Recommendation is foreseen to be one of the most important services that can provide such personalized multimedia contents to users\cite{ZWZ}. Several companies have demonstrated initial successes in multimedia recommendation system design. \cite{youtube} reported that YouTube won its first Emmy for video recommendations. Actually, most OSNs recommend video content to their users based on the user's rich context information (e.g., social
status, ages, professions, health conditions and hobbies) contained in their released multimedia data. Regarding this way, several recommendation systems have been proposed \cite{CEM},\cite{LHL}.




However, there exist two major challenges in this scenario. The first challenge comes from the big data's role in the personalized recommendation. In detail, OSNs have accelerated the popularity of applications and services, resulting in the explosive increase of social multimedia data. In this case, multimedia big data puts companies in a favorite position to have access to much more contextual information\cite{MCH}. However, how to harness and actually use big data to effectively personalize recommendation is a monumental task. Traditional stand-alone multimedia systems cannot handle the storage and processing of this large-scale datasets\cite{ASM}. Besides that, complex and various user-generated multimedia big data in the OSNs results in the sparsity and heterogeneity of users' context data. Hence, it is extremely challenging to implement recommendation with the multimedia big data.

Furthermore, the privacy in recommendation has raised widely concern. On the one hand, as declared in \cite{AJM}, user's sensitive context information may be exposed by the recommendation results.
Intuitively, the more detailed the information
related to the user is, the more accurate the recommendations for the user are. But once the recommendation records are accessed by a malicious third party, individual features can be inferred by them merely based on the outcome of the recommendation. For example, advertising video of luxury goods recommended to a particular person indicate the income level of this user. Also basketball video recommendation for the same user exposed it's hobby. Then  with additional side information, the malicious party may identify the person in real life.
On the other hand, the inventory of videos is an important commercial secret for the service vendor. As for the service vendors' incentives, they rely on stored video source files to gain popularity among users. Intuitively, video service vendors are selfish and  they refuse the inference of what they have in the inventories by the revenue gain of each video. Consequently, avoiding the divulge of video contents of each service vendor is desirable.

Taking the above two difficulties into consideration, establishing a privacy-preserving video recommendation system with multimedia bid data can be extremely challenging. Traditional recommender systems for multimedia, including collaborative filtering (CF)\cite{SJX} and content-based (CB) recommendation\cite{MTM} 
can provide meaningful multimedia recommendations at an individual level. However, their stand-alone systems have difficulties in dealing with tremendous high-dimensional multimedia big data. As for the privacy concern in recommendation, previously, anonymity was the main tool in recommendation\cite{SS}. But the fact that the information can only be partially removed will allow for re-identification.


Differential privacy \cite{CD} proposed recently is a heuristic method to solve this problem. Informally, differential privacy means that the output is going to be almost exactly the same whether it includes a single user's data in the input datasets. Therefore, hardly can one make an accurate inference on signal user's feature based on the recommendation results. Besides, adding laplace noise into the recommendation rewards can hide small changes that arise from a single video's contribution. Thus, the revenue gain of one signal video cannot be deduced. Several studies have incorporated it into recommendation systems \cite{ZJ, MCS}, but their works only focus on small-scale media datasets, yet executing differential privacy in a large datasets often impacts little on accuracy, which works extremely efficiently under the big data context In conclusion, it is necessary to design a privacy-preserving video recommendation that can handle the multimedia big data and achieve high-accurate recommendations.

In this paper, we introduce differential privacy into distributed online learning to design an efficient and high-accurate timely recommendation system based on multimedia cloud computing\cite{WZC}. As illustrated in Fig. 1, user-generated multimedia big data (e.g, images, audio clips and videos) is first translated to remote media cloud and stored in decentralized data centers (DCs). Then use technologies such like Bag-of-Features Tagging (BoFT)\cite{MC} to extract user's context vectors and convert the results to distributed video service vendors (servers). Finally recommended video contents are pushed to multimedia applications in OSNs.

Our main theme in this media cloud based scenario is that video service vendors are modeled as decentralized online learners, who try to learn from user's high-dimensional context data and match it to the optimal video. The service vendors are connected together via a fixed network over the media cloud, each of whom experience inflows of users' context vectors to them. If service vendors cannot find suitable videos in their repositories for the coming user, they can forward the use's context data to neighbor service vendors, who will find out the suitable video in his repository to recommend to this user. At the end of each time slot, the reward of the recommended video is observed. Service vendors can learn from the result and adjust their selection strategy next time. Since the extracted context vectors from multimedia big data are high-dimensional and omnifarious, the context space with $d$ dimensions ($d$ is the number of user features) can be extremely huge and heterogeneous. Then, learning the most matchable video for each individual can be extremely slow. Therefore, each service vendor initially groups users (partition the context space) with similar context into rough crowds,
   \begin{figure}[!htb]
 \vspace{-.1em}
\centering
\includegraphics[scale=.58]{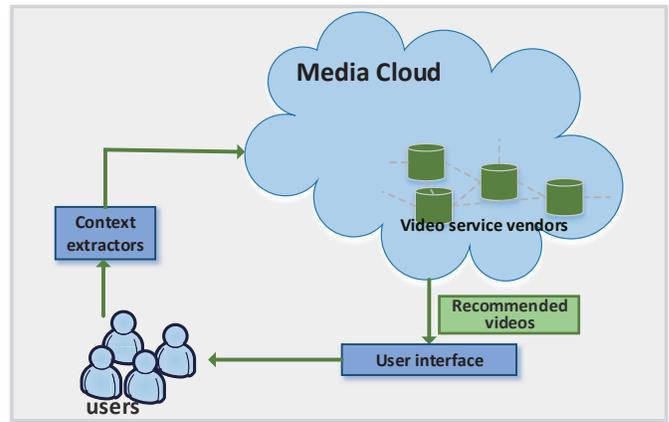}
\caption{A general illustration of multimedia cloud based video recommendation system.}
\label{fig:digraph}
\vspace{0.1mm}
\end{figure}
and then they dynamically refine the partition strategies over time.

To goal of each service vendor is to maximize its long term expected total recommendation reward and do not want to reveal their repositories to other service vendors. However, in the cooperation, each service vendor will share some information such as the user's context vectors and the videos' revenue gains with neighbor service vendors. Then, service vendors can infer the repositories of other service vendors from the shared information. To solve this privacy leakage, we adopt \emph{Laplace mechanism}\cite{CD}, adding noise to shared revenue gains. As for the users' privacy, to prevent the exposure of their feature by the recommendation videos, adding noise to the revenue gains is not noneffective. Because the gain is produced after the recommended video is revealed and disturbing the accurate estimation of gains of their own videos with this noise is not necessary. Thus, we employ \emph{exponential mechanism}\cite{FM} to protect the users' privacy, where the service vendors randomly select the video according a computed exponential probabilities.
Faced with the fact that user's contexts ($d$-dimensional point in the context space) are sparse distributed over the context space,
we propose a novel \emph{geometric differentially private} method to promote the total reward. This paper makes the following contributions:
\begin{itemize}
\item We propose a media cloud based video recommendation system and rigorously formulate it as a distributed online learning problem. In our model, decentralized service vendors work cooperatively to deal with large-scale contextual data.

\item To handle the dimensionality and sparsity of the multimedia big data, our method adaptively partitions the context space for each service vendor. Our evaluation results show this method has lower performance loss and converges fast to optimal strategy.

\item To the best of our knowledge, we are the first to deal with the privacy issue of both the social media users and video service vendors in recommendation. We integrate exponential mechanism and Laplace mechanism simultaneously into distributed learning systems. We guarantee $\varepsilon $-differential privacy while not coming at substantial expense in total reward.

\item We propose a ``geometric differentially private model'' to deal with the sparse contextual data, which can reduce the performance loss extensively.
\end{itemize}

The remainder of the paper is organized as follows. In Section II, we briefly review the related work. Section III presents the necessary background concepts of this work. In Section IV, we detail the system model, define our performance metric, the adversary model and design goals. Section V describes the design of algorithms and provides theoretical analysis of the performances. In section VI, we present our geometric differentially private model. Section VII discusses our experimental results and analysis. Section VIII concludes this paper.
\section{Related Work}

Several recommendation algorithms have been exploited in the past. Content-based filtering (CB) recommendation systems \cite{MJ, EG, LL} focus on the similarities of content titles, tags and descriptions and they find user-interested items based on user's individual reading history. CB recommender systems are easy to deploy. Nonetheless, simply representing the user¡¯s profile information by a bag of words is not sufficient to capture the exact interests of the user. Collaborative filtering (CF) recommendation systems \cite{GG, ZNC} rely on abundant user transaction histories and content popularity. CF systems require enough history consumption record and feedback, which is not suitable to real-time recommendation. Graph-based (GB) recommendation systems \cite{ZW, SB} build a graph to calculate the correlation between recommendation objects. Then, recommendation problem turns into a node selection problem on a graph. Besides that, users¡¯ cotagging behaviors and friendships in social network are described by a graph. Combining graph theory with recommendation is a marvellous idea. However, in OSNs, this graph can be continuously changeable. Constructing and storing such graph are impractical. Context-aware recommendation systems make recommendation based on the contextual information both of items and users. \cite{LHL} has done a pioneering in this area, but its centralized framework fails to satisfy the need of big data environment. Our distributed cooperative recommendation framework can arrange recommendation timely under big data environment and provides rigorous performance guarantees.

As for the privacy in recommendation systems, anonymity was the main tool\cite{SS}. However, especially for rich, high-dimensional big data, most anonymization techniques appear to cripple the utility of the data\cite{AC, BJ}. In addition, though anonymized, users may be re-identification in the presence of colluding adversaries or those with auxiliary information\cite{ANV}. On the other hand,  prior works lay emphasis on cryptography\cite{Erkin} to make the privacy-sensitive data inaccessible to any outsiders and the server by means of encryption. But it usually incurs high computation and communication overheads. Differential privacy \cite{CD} proposed in recent years has been incorporated into recommendation by several studies. McSherry and Mironov \cite{MCS} show how to adapt the leading algorithms used in the Netflix Prize competition to make privacy-preserving recommendations. This is typically accomplished by adding noise to the item covariance matrix, to hide small changes that arise from a single user¡¯s contribution.  Ashwin \emph{et al.}\cite{AM} and Jorgensen\cite{ZJ} combine differential privacy with social graph
for recommendation. But their work only study the privacy of sensitive user-item preferences and connections between people, rather than individual features. Our work aims at the privacy of individual features contained in their context data and the secrecy of service vendors' data.
 \begin{table}
 \vspace{-.5em}
\centering
\renewcommand{\multirowsetup}{\centering}
\caption{Comparison with prior work in recommender
systems.}
\begin{tabular}{|c|c|c|c|c|}
\hline
\multirow{5}{0.4cm}{} & \multirow{5}{2.1cm}{Content-based\\(CB),Collaborative filtering(CF),graph-based(GB),context-aware(CA)}
& \multirow{5}{1.7cm}{Anonymity(A),\\cryptography\\(Cr),differential privacy(DP)} & \multirow{5}{0.7cm}{Private target}
& \multirow{5}{1.0cm}{Centralized\\(C), decentralized\\(D)}\\
&&&&\\
&&&&\\
&&&&\\
&&&&\\
\hline
\cite{MJ, EG, LL} & CB & $\backslash $ & None & C \\
\hline
\cite{GG, ZNC} & CF & $\backslash $ & None & C \\
\hline
\cite{ZW, SB} &	GB & $\backslash $ & None & C \\
\hline
\cite{AC} & CA & $\backslash $ & None &	C \\
\hline
\cite{CEM} & CA & $\backslash $ & None &	D \\
\hline
\cite{SS} & Hybrid CF & A & User & C \\
\hline
\cite{Erkin} & CF & Cr & User & C \\
\hline
\cite{ZJ} & GB & DP & User & D \\
\hline
\cite{MCS} & CF & DP & User & C \\
\hline
\cite{AM}  & GB & DP & User &C \\
\hline
\multirow{3}{0.4cm}{Our work} & \multirow{3}{2.1cm}{CA} & \multirow{3}{1.7cm}{DP} & \multirow{3}{0.8cm}{User,\\service provider} & \multirow{3}{1.2cm}{D} \\
&&&&\\
&&&&\\
\hline
\end{tabular}
\end{table}

\section{Backgrounds}

\subsection{Differential Privacy}
The concept of \emph{differential privacy} is originally introduced by Dwork\cite{CD}, which gives us a riorous definition of privacy.\vspace{1mm}

\textbf{Definition 1} \emph{
(Differential Privacy \emph{\cite{CD}}). A randomized algorithm
M has $\varepsilon $ differential privacy if for any two input sets
$A$ and $B$ with a single input difference, and for any set of
outcomes $R \in Range(M)$,}
\[\mathbb{P} [M(A) \in R] \le \exp (\varepsilon ) \times \mathbb{P} [M(B) \in R].\]

Informally, differential privacy means that the outcome of two
nearly identical input datasets (different for a single component)
should also be nearly identical. Thus, attacker is not able to get the information of the  individual's information by comparing the query result of A and B. In our model, the input datasets are users' context vectors. The privacy $\varepsilon$ is the parameter to measure the privacy level of the algorithm. The choice of $\varepsilon$ is a trade-off between the privacy and the accuracy of the output.

One effective tool is the Laplace Mechanism\cite{CD}, i.e., $M(x) = f(x) + Lap(\frac{{\Delta f}}{\varepsilon })$. In this way, $f()$ is a counting query on the data set $X$, and $Lap()$ is the Laplace distribution with standard deviation $\frac{{\sqrt 2 \Delta f}}{\varepsilon }$ to scale the counting query result.


\textbf{Definition 2} \emph{
(Sensitivity of Laplace mechanism\emph{\cite{book}}). The sensitivity of a function $f$ is:}
\begin{equation}
\Delta f = \mathop {\max }\limits_{x,y} {\left\| {f(x) - f(y)} \right\|_1},
\end{equation}
where $x$ and $y$ are input datasets differ on at most one component. The sensitivity of a function $f$ captures the magnitude, by which
a single component can change the function $f$ in the worst case.
Indeed, the sensitivity of a function
gives an upper bound on how much we must perturb its output to preserve privacy.\vspace{1ex}

\textbf{Corollary 1} \emph{
(Composability \emph{\cite{FM}}). The sequential application
of randomized computation ${M_i}$, each giving ${\varepsilon _{\rm{i}}}$ differential
privacy, yields $\sum\nolimits_i {{\varepsilon _{\rm{i}}}} $ differential privacy.}\vspace{1ex}

Referring to differential privacy, another powerful tool is the exponential mechanism \cite{FM}. The exponential mechanism ${M_{E(x,u,R)}}$ selects and outputs an
element $r \in R$ with
probability proportional to $\exp (\frac{{\varepsilon u(x,r)}}{{2\Delta u}})$.  Here, $x$ is the input data set we want to protect, $r$ is the output of the mechanism and $u(x,k)$ is the unity function. There is also a definition of the sensitivity:\vspace{1ex}

\textbf{Definition 3}\emph{
(Sensitivity of Exponential Mechanism \emph{\cite{book}}). The sensitivity of exponential mechanism is defined as follows:}
\begin{equation}
\Delta u  = \mathop {\max }\limits_{r \in R} \mathop {\max }\limits_{{x_1},{x_2}:{{\left\| {{x_1} - {x_2}} \right\|}_1} \le 1} \left| {u({x_1},r) - u({x_2},r)} \right|.
\end{equation}

The sensitivity measures the change of utility function ${u({x},r)}$, when one item in targeted data set changes. An important theorem can also be derived as \cite{book} :\vspace{1ex}

\textbf{Theorem 1.} \emph{Fixing a database $x$, let ${R_{OPT}} = \left\{ {r \in R:{\rm{u}}(x,r) = OP{T_{\rm{u}}}(x)} \right\}$ denote the set of elements in R which attain utility score $OP{T_u}(x)$. Then, When used to select an output $r \in R$, the exponential mechanism $\varepsilon _q^\varepsilon (x)$ ensures that:}
\begin{equation}
\begin{array}{l}
\mathbb{P} [u(x,\varepsilon _q^\varepsilon (x) ) < \mathop {\max }\limits_r u(x,r) - \frac{{2\Delta u}}{\varepsilon }(\ln (\frac{{\left| R \right|}}{{\left| {{R_{OPT}}} \right|}}) + t)]\\
 \ \ \ \le \exp ( - t).
\end{array}
\end{equation}

\subsection{Online Learning}
 Our proposed distributed learning method derives from contextual bandits\cite{JL}. This algorithm learns form the context information available at each time, which, in this case, is the users' context vectors. Then, it keeps an index that weights the \emph{estimated performance and uncertainty} of each action (recommended video or neighbor service vendor in this case) and choose the action with highest index at each time. Furthermore, the indices for the next time slot for all actions are updated based on the feedback received from the chosen action (user¡¯s click feedback). There exist some works studying the contextual bandit \cite{JL, AS}, where the best action given the context is learned online.
C. Tekin \emph{et al.} first proposed a distributed contextual bandit framework for big data classification \cite{cem} and social  recommendations \cite{CEM}. But the uniform partition method proposed in their work does not fit into the sparse big data. A context-aware partition method for big data proposed in \cite{JX} is a heuristic work. Nonetheless, the single-learner framework can not satisfy the need of the massive big social data. We combine adaptive context space partition with distributed learning, which can efficiently handle above difficulties.

  \begin{figure}[!htb]
 \vspace{-.1em}
\centering
\includegraphics[scale=.42]{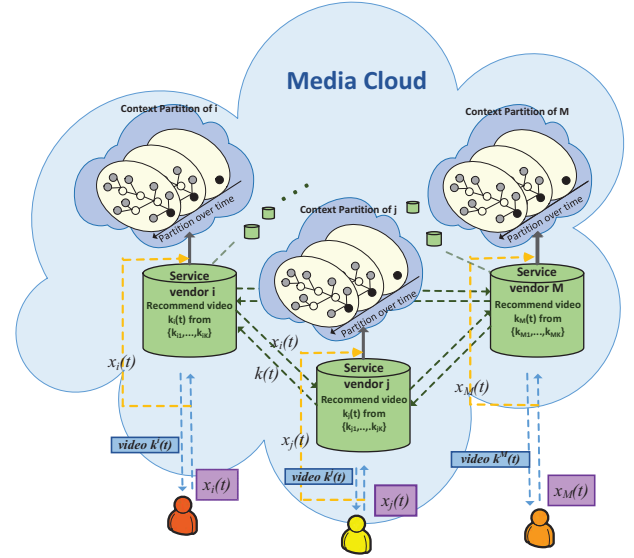}
\caption{A general explanation of our video recommendation system. Each service vendor keeps a context space partition of arriving contexts. This partition process is dynamic by time.}
\label{fig:digraph}
\vspace{-1em}
\end{figure}

\section{Problem Formulation}
In this section, we first present the system model and assumptions. Then we give our performance metric. Finally, we outline the adversary model and design goals.

\subsection{System Model}
The system model is shown in Fig. 2. There are $M$ distributed service vendors distributed in media cloud, which
are indexed by set $\mathcal{M}=\{1,2,3...,M\}$. They work independently and cooperatively in
discrete time setting $t = 1,2,...,T$. Each vendor owns a set of videos. We denote the set of videos ${{\cal M}_i} = \left\{ {{k_{1,}}{k_{2, \cdots ,}}{k_K}} \right\}$ for service vendor $i$. At each time slot, the following
events happen sequentially  for service vendor $i$: 1) a user's extracted context vector ${x_i}(t)$ comes to service vendor $i$; 2) The
service vendor $i$  chooses one video from his repository ${{\cal M}_i}$ or sends the context vector to neighbor service vendor $j$, who will select one video from ${{\cal M}_j}$ for the user with this context; 3) At the end of each time slot, the user's click feedback ${f_{k,{x_i}(t)}}(t)$ (If user clicks, it equals one, otherwise zero, where $k$ is the recommended video.) is observed; 4) The
service vendor $i$ learns from the feedback, then promotes the selection strategy for next user.


We describe the details and some reasonable assumptions here.

1) Each service vendor has access to only its own video repository. Service vendors are selfish in the sense that, they do not reveal their repositories to other service vendors. But they know the number of videos of other service vendors.  In this article, we assume every service vendor possesses $K$ videos.

2) The context information ${x_i}(t)$
of the data is a high-dimensional vector. Each coordinate of the vector represents the
feature of the user (e.g., gender, hobby, profession and age). We use the hypercube $\mathcal{X}  = {\left[ {0,1} \right]^d}$ to denote its range, where $d$ is the dimension of the space.  Given the setting of big data,  $d$ is extremely large and those context vectors are distributed non-uniformed in the hypercube space.

3) At the end of each time slot, we use a random variable ${f_{k,x}}({\rm{t}})$ to represent the reward (user click feedback) produced by the recommended video $k$. If user clicks the recommended video $k$, it equals one, otherwise zero. Let ${u_{k,x}}$ be the expected reward of a video conditional on the context $x$. Different videos have different expected reward for the same context. We aim to find the video with the highest expected reward for that context. Naturally, similar contexts have similar expected reward with the same video. We use the Lipschitz condition to describe this similarity:
\begin{equation}
\left| {{u_{k,{x_1}}} - {u_{k,{x_2}}}} \right| \le L{\left\| {\left. {{x_1} - {x_2}} \right\|} \right.^\alpha }.
\end{equation}

The goal of the service vendor is to try its best to recommend video with highest expected reward. Consequently, if the service vendor
does not have matchable video to its coming user's context, he will forward the context to neighbor service vendor. Our algorithm chooses
another service vendor by comparing the average rewards of each service vendor with those of its own videos. To be reasonable, in this distributed contextual bandit framework, we call ${{\cal K}_i} = {{\cal M}_i} \cap {{\cal M}_{-i}}$ the set of arms (videos and other service vendors) of service vendor $i$, where $ {{\cal M}_{-i}}= \mathcal{M} - \{ i\} $.

\subsection{Performance Metric}

 \textbf{Definition 4}\emph{
(Optimal Arm). Our benchmark when evaluating the performance of the learning algorithm is the
optimal solution, which selects the arm $k$ with the highest expected reward from the set ${{\cal K}_i} = {{\cal M}_i} \cap {{\cal M}_{-i}}$ given context ${x_t}$ at time
$t$.  Specifically, the optimal arm we compare against is given by:}
\begin{equation}
{k^*}\left( {{x_t}} \right) = \mathop {\arg \max }\limits_{k \in {{\cal K}_i}} {u_{k,{x_t}}},\forall {x_t} \in \mathcal{X}.
\end{equation}

Knowing the optimal solution means that learner $i$ (service vendor $i$ in this case) knows the arm in ${{\cal K}_i}$ that yields the highest expected accuracy for each $x_t \in \mathcal{X} $.

\textbf{Definition 5} \emph{
(The Regret of Learning). We define the regret as a performance measure of the learning
algorithm used by the learners.
Simply, the regret of a learning algorithm for
learner $i$ is the reward gap between optimal arms and selected arms:}
\begin{align}
R(T) = \sum\nolimits_{t = 1}^T {{{\rm{u}}_{{k^*}({x_t}),{x_t}}} - E\left[ {\sum\nolimits_{t = 1}^T {{f_{k({\rm{t}}),{x_t}}}(t)} } \right]},
\end{align}
where $k(t)$ denotes the video or neighbor service vendor chosen at time $t$, ${k^*}({x_t})$
denotes the best choise for context ${{x_t}}$. Regret gives the
convergence rate of the total expected reward of the learning algorithm to the value of the
optimal solution.

\subsection{Adversary Model and Design Goals}
As similar privacy concern for the users' sensitive context data in \cite{AJM}, we consider a adversary model as follows: (1) \emph{Malicious third party} who can gain access to the recommendation outputs and own some side information such as location about some users. The goal of this malicious third part is to deduce a particular user's features by observing the recommendation outputs. Then, they can identify the media user in the real world with deduced features and additional side information. (2) \emph{Selfish and curious service vendors} who want to infer neighbors' repositories from shared information. For example, the curious service vendor forward a sports fan's context to a neighbor service vendor, who output a video and receive high reward. Then, the curious service vendors know that this neighbor service vendor owns a video about sport.

To address the adversary models above, we proposed a differentially private learning algorithm. Our scheme achieves privacy protection and performance guarantees as follows:
\begin{itemize}
\item \emph{Users' Privacy Guarantee}: Even if the malicious party can gain access to the recommendation outputs, it is less likely for he to infer the user's feature from the recommended result. And we prove that our proposed algorithm can preserve $\varepsilon $-differential privacy for user's privacy.
\item \emph{Service vendors' Privacy Guarantee}: The curious service vendor can not distinguish the video of neighbor service vendors by shared information. The proposed algorithm can preserve $\varepsilon $-differential privacy for service vendors.
\item \emph{Performance Guarantee}: Our proposed algorithm can guarantee the regret in equation (6) is sublinear converged, i.e., $ R(T) = O({T^\gamma })$ such that $\gamma
< 1$ . A smaller $\gamma $ will result in faster convergent rate. In the following section we will propose a private
distributed learning algorithm
with sublinear regret.
\item \emph{Privacy-Reward Trad-off}: Our analysis shows that the higher level the privacy is preserved, the lower the total reward is received. By varying the value of the privacy parameter $\varepsilon $, we can keep a trad-off between the total recommendation reward and the privacy preservation level.
\end{itemize}

\section{Differential Private Distributed Online Learning Algorithm for Cloud Based Video Recommendaion }
Since the reward of each recommended video for different users have unknown stochastic distributions, the natural way to learn a video's performance is to record and update its sample mean reward for the same context vector. Using such an empirical value to evaluate the expected reward is the basic approach to help the service vendors to learn.
However, the context space $\mathcal{X}$ can be very large, recording and updating the sample mean reward for each context are scarcely possible. The memory capacity of the sever can not meet the need of keeping a
sample mean reward for all contexts. To overcome the difficulty, we dynamically partition the entire context space into multiple smaller context subspaces (according to the number of arriving users). Then, we maintain and update the sample mean reward estimates for each subspace. This is due to the fact that the expected rewards of a video are likely to be similar for similar contexts.

In our distributed framework, each service vendor $i \in \mathcal{M}$ dynamically partitions the context space $\mathcal{X}$ when context ${x_i}(t)$ arrives to them. To better understand the proposed P-DAP algorithm, we apart it into two algorithms, i.e., Algorithm 1 and Algorithm 2. Service vendor $i$ runs Algorithm 1 to select video or request neighbor service vendor's help for its own user. Because service vendor $i$ does not outward recommendation revenue gain to other service vendors, we only need to protect user's privacy and we adopt exponential mechanism in Algorithm 1 (named as ExP-DAP) to achieve this protection.
When service vendor $i$  receives users' extracted context vectors forwarded from other service vendors, it runs Algorithm 2 (named as LaP-DAP) to select videos and protect the privacy of selected videos.
Two algorithms are carried out simultaneously, although we describe them separately.

Next we present our online learning algorithm. In section VI, we will refine the proposed algorithm to \emph{geometric differential privacy} to reduce the performance loss.

 \subsection{Algorithm Description}
 In this subsection, we describe our differentially \underline{P}rivate \underline{D}istributed learning with \underline{A}daptive context space \underline{P}artition algorithm (P-DAP for short) for video recommendation.  We first introduce several useful concepts for describing the proposed algorithm.

\begin{itemize}
\item \textbf{Context subspace.} A context subspace $C$ is a subspace of
the entire context space $\mathcal{X}$, i.e., $C \subseteq \mathcal{X} $. In this paper,
all context subspaces are created by uniformly
partitioning the context space on each dimension.
Thus, each context subspace is a $d$-dimensional hypercube
with side length being ${m^{ - l}}$, where $m$ is number of segmentations of each dimension to be partitioned and $l$ is the partition level. To be specific, when we assign $m = 2$, $d = 1$ and entire
space is $[0,1]$,
then the entire context space $[0,1]$ is a level-0 subspace, $[0,{1 \mathord{\left/
 {\vphantom {1 {2)}}} \right.
 \kern-\nulldelimiterspace} {2)}}$
and $[{1 \mathord{\left/
 {\vphantom {1 2}} \right.
 \kern-\nulldelimiterspace} 2},1]$ are two level-1 subspaces etc.

\item \textbf{Active context subspace.} We define a set named ${P^t}$ in which all existing subspaces is collected, and ${P^t}$ is changing over time. For example, when $d = 1$, $\{ [0,1]\} ,\{ [0,1/2),(1/2,1]\} $ are two sets of active context subspaces. A context subspace $C$ is active
if it is in the current context subspace set ${P^t}$, i.e. $C \in {P^t}$.

\item \textbf{Notations.} For service vendor $i$ and each active context subspace $C \in {P^t}$, the algorithm
maintains a counter $N_{k,C}^i(t)$ recording the number of times when $k$ is selected for contexts belong to subspace $C$. ${\overline r^i _{k,C}}(t)$ estimates the sample mean reward of video $k$ for the context subspace $C$ up to time $t$. We have $\bar r_{k,C}^i(t) = \sum\nolimits_{x(t) \in C} {f_{k,x(t)}^i/N_{k,C}^i(t)} $. The algorithm also maintains a counter $M_C^i(t)$ that records the number of context arrivals to $C$ up to time $t$.
\end{itemize}
\begin{algorithm}
\caption{ExP-DAP for service vendor $i$'s own user}
\begin{algorithmic}[1]
\STATE Input: ${{k}} \in {{\cal K}_i}$; $m$, $p$, $A$, $K$, $\epsilon$, $\Delta u$, ${G_1}(t)$, ${G_2}(t)$, ${G_3}(t)$.
\STATE Initialize: ${P^t} = \{ {\cal X}\}$, $\bar r_{k,C}^i(0) = 0$, $\forall k \in {{\cal K}_i}$,
$M_C^i\left( 0 \right) = 0$, $N_{k,C}^i(0) = 0$, $l = 0$
\FOR {$t = 1,...,T$, ${x_i}(t) \in C$ }
 \IF{$\exists k \in {{\cal M}_i}$, such that $N_{k,C}^i(t) < {G_1}(t)$ }
   \STATE  Select $k$ and observe $ f_{k,C}^i(t)$.
 \ELSIF{$\exists k \in {{\cal M}_{-i}}$, such that $N_{1,k,C}^i(t) < K{G_3}(t)$}
  \STATE Forward ${x_i}(t)$ to service vendor $k$.
  \ELSIF{$\exists k \in {{\cal M}_{-i}}$, such that $N_{k,C}^i(t) < {G_2}(t)$ }
  \STATE Forward ${x_i}(t)$ to service vendor $k$ and receive $ f_{k,C}^i(t)$.
\ELSE
\FOR{all $k \in {{\cal M}_i} $}
\STATE \begin{small}$\mathbb{P}[select\ k] = {{\exp \left( {\frac{{\epsilon \overline r_{k,C}^i\left( t \right)}}{{2\Delta u}}} \right)} \mathord{\left/
 {\vphantom {{\exp \left( {\frac{{r_{k,C}^i\left( t \right)}}{{2\Delta u}}} \right)} {\sum\nolimits_{k \in M{\rm{i}}} {\exp \left( {\frac{{r_{k,C}^i\left( t \right)}}{{2\Delta u}}} \right)} }}} \right.
 \kern-\nulldelimiterspace} {\sum\limits_{ k \in {{\cal M}_i}} {\exp \left( {\frac{{\epsilon \overline r_{k,C}^i\left( t \right)}}{{2\Delta u}}} \right)} }}.$\end{small}
\ENDFOR
\STATE Select $k_i \in {{\cal M}_i} $ according to computed probability distribution.
\STATE Select ${k_j} \in {{\cal M}_{-i}}$ such that ${k_j} = \mathop {\arg \max }\limits_{k \in {{\cal M}_{-i}}} \overline r _{k,C}^i(t)$.
\STATE Call $k$ such that $\overline r _{k,C}^i(t) = max\left( {\overline r _{{k_i},C}^i(t),\overline r _{{k_j},C}^i(t)} \right)$.
\ENDIF
\STATE Update $M_C^i\left( t \right)$, $N_{k,C}^i(t)$, $\bar r_{k,C}^i(t)$.
\IF{$M_C^t \ge A{m^{pl}}$}
\STATE Partition $C$.
\ENDIF
\ENDFOR
\end{algorithmic}
\end{algorithm}

To begin with, we present our Algorithm 1 in the following 3 phases:
  \begin{figure}[!htb]
 \vspace{-.5em}
\centering
\includegraphics[scale=.50]{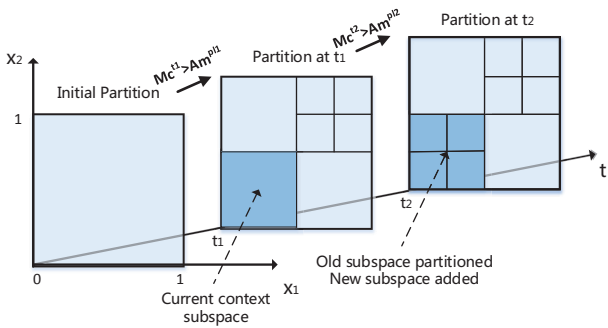}
\caption{A process of dynamic partition of context space}
\label{fig:digraph}
\vspace{-.5em}
\end{figure}
\noindent \textbf{Phase 1: Exploration and Reward Estimation}

Upon each context data arrival, service vendor $i$ first checks to which subspace $C$ in the set ${P^t}$ the context belongs and the level of $C$. To get accurate performance estimation of each arm $k \in {{\cal M}_i}$, service vendor $i$ needs to judge whether $k$ has been fully explored (line 4, 5). Since service vendor $i$ does not know the performance service vendor $k$'s videos, it needs to send neighbor service vendor $k$ some context samples to train it and make sure it will mostly select optimal video. The $N_{1,k,C}^i(t)$ denotes the times when $k \in {{\cal M}_{-i}}$ is selected for training. In the training process, service vendor $i$ dose not need to communicate with service vendor $k$ to observe the reward $f_{k,x(t)}^i(t)$ (line 6, 7). If each service vendor $k \in {{\cal M}_{-i}}$ has been fully trained, service vendor $i$ start to explore the performance of leaner $k \in {{\cal M}_{-i}}$ and observe the reward of each $k$ (line 8, 9). The control function ${G_1}(t)$, ${G_2}(t)$ and ${G_3}(t)$ ensure that video is selected sufficiently many number of times so that the sample mean estimates $\overline r _{k,C}^i(t)$ are accurate enough. And we set different control function for $k \in {{\cal M}_{-i}}$ and $k \in {{\cal M}_{i}}$, i.e., ${G_2}(t)$ is larger than ${G_1}(t)$. Because for $k \in {{\cal M}_{-i}}$, the reward $\overline r _{k,C}^i(t)$ is added with noise, we need more times to evaluate performance of $k \in {{\cal M}_{-i}}$.

%

\noindent \textbf{Phase 2: Decision with Privacy Protection}

For subspace $C$, when all arms have been fully explored, there are accurate sample mean estimations for each arm. In traditional bandit algorithms, the learners (service vendor in this case) usually select the arm with the highest sample mean reward. However, the optimal arm will expose the individual feature. Thus, to protect the user's privacy, service vendor $i$ first randomly choose one arm $k_i \in {{\cal M}_i} $ according to the computed probability distribution, where $\Delta u$ is the sensitivity of exponential mechanism (line 11-14). Then, it select another arm $k_j \in {{\cal M}_{-i}} $ with the highest estimated reward. Finally service vendor $i$ compare the estimated reward of $k_j$ and $k_i$, then it select the one with higher estimated reward for context ${x_i}(t)$ (line 15, 16). We will prove this randomly selection scenario guarantee $\varepsilon $-differential privacy in next our analysis section.\vspace{1ex}


\noindent \textbf{Phase 3: Update and Partition the Context Subspace}

At the end of each time slot, the algorithm first updates $M_C^i\left( t \right)$, $\bar r_{k,C}^i(t)$ and $N_{k,C}^i(t)$, where $M_C^i\left( t \right)=M_C^i\left( t \right)+1$, $N_{k,C}^i(t)=N_{k,C}^i(t)+1$ and $\bar r_{k,C}^i(t) = \sum\nolimits_{x(t) \in C} {f_{k,x(t)}^i/N_{k,C}^i(t)} $. Then
 the algorithm decides whether to further partition the current subspace $C$, depending on whether we have sufficient context vectors arrivals in $C$. Specifically, if $M_C^i(t) \ge A{m^{pl}}$ at time $t$, $C$ will be further partitioned, where $p$ and $m$ are positive numbers. When partitioning is needed, $C$ is uniformly partitioned into ${m^d}$ smaller hypercubes. Each hypercube is a level-$(l+1)$ subspace with side-length $1/m$ of that of $C$. Then $C$ is removed from the current context set ${P^t}$. New subspaces are added into ${P^t}$. Fig. 3 provides us an illustration of this partition process when $m = 2$, $d = 2$.


Then, we describe Algorithm 2 as follows. In our problem setting, in order to protect the privacy of neighbor service vendors, we face a big challenge that traditional differential privacy only apply to static database. By contrast, the datasets we want to protect are dynamically releasing over time. In detail, suppose at every time step ${\rm{t}} \in [T]$, one entry from dataset $D$, ${f_{k,x(t)}} \in \{ 0,1\} $ arrives and the task is to output ${v_t} = \sum\nolimits_{\tau  = 1}^t {{f_{k,x(\tau )}}} $ while ensuring the complete output sequence $\left\langle {{v_1},...,{v_T}} \right\rangle $ is $\varepsilon $-differential private. To overcome this challenge, we use a \emph{tree based aggregation} method initially proposed by Dwork \cite{d4}, Chan  \cite{d7}.

 \textbf{Tree based aggregation.} Assume for simplicity that $T = {2^\alpha }$ for some positive integer $\alpha $. We create a binary tree, i.e., $Tre{e_k}$ for each video $k \in {{\cal M}_{i}}$ with its leaf nodes being ${f_1},...,{f_T}$. As illustrated in Fig. 4, at each time slot, when new reward is produced, we insert the value of the reward into the leaf node. Over the entire time sequence $[T]$, the rewards are inserted sequentially. Each internal node $x$ in
$Tre{e_k}$ stores the sum of all the leaf nodes in the tree rooted at $x$. First notice that one can compute any
${v_t}$ using at most $\log (T)$ nodes of $Tre{e_k}$. Second, notice that for any two neighboring datasets $D$ and
$D'$ different in leaf node ${f_i}$ and ${f_i}'$  at most $\log (T)$ nodes in $Tre{e_k}$ gets modified. So, if we flatten the complete tree as a vector then
for any neighboring datasets $D$ and $D'$ one can easily show that ${\left\| {Tree(D) - Tree(D')} \right\|_1} \le \log (T)$. We will further bound the amount of the noise added to each tree in section V when evaluating the performance of our algorithm.\vspace{1ex}
\begin{algorithm}
\caption{LaP-DAP for other service vendors' users}
\begin{algorithmic}[1]
\STATE Input: ${{k}} \in {{\cal M}_i}$; $m$, $p$, $A$, $T$, $\epsilon$, ${G_3}(t)$.
\STATE Initialize ${P^t} = \{ {\cal X}\}$, $l = 0$, $M_C^i\left( 0 \right) = 0$, $\forall k \in {{\cal M}_i}$, $\bar r_{k,C}^i(0) = 0$, $N_{k,C}^i(0) = 0$, $F_{{k},{x_i}(t)}^i(0) =0$
\STATE Create empty binary tree $Tre{e_k}$ with $T$-leaves, $\forall k \in {{\cal M}_i} $.\vspace{-3ex}
\FOR {$t = 1,...,T$, ${x_j}(t) \in C$}
 \IF{$\exists k \in {{\cal M}_i}$, such that $N_{k,C}^i < {G_3}(t)$ }
   \STATE  Select $k$ and insert observed $ f_{k,C}^i(t)$.
 \ELSE
 \STATE  \small{Select ${k^*} = \mathop {\arg \max }\limits_{k \in {{\cal M}_i}} \overline r _{k,C}^i\left( {\rm{t}} \right)$ and observe $f_{k^*,{x_j}(t)}^i(t)$.}
 \STATE Insert $F_{{k^*},{x_j}(t)}^i(t) = f_{{k^*},{x_j}(t)}^i(t) + Lap(\frac{{\Delta f}}{\varepsilon })$ to $Tre{e_{{k^*}}}$
\ENDIF

\STATE Update $M_C^i\left( t \right)$, $\bar r_{k,C}^i(t)$, $N_{k,C}^i(t)$.
\IF{$M_C^t \ge A{m^{pl}}$}
\STATE Partition $C$.
\ENDIF
\ENDFOR
\end{algorithmic}
\end{algorithm}

\textbf{LaP-DAP Description.} When service vendor $i$ receives context ${x_j}(t)$ from  service vendor $j$, service vendor $i$ first determines the subspace $C$ to which this context belongs and the level $l$ of it. Then we want to make sure whether each video $k \in {{\cal M}_i}$ has been selected for enough times for accurate estimation (line4-6). If each video has been explored sufficiently, we select the video ${k^*}$ with highest accuracy and observed the reward $f_{k^*,{x_i}(t)}^i(t)$. Because after the training process, service vendor $j$ can gain access to this observed reward of service vendor $i$ and make evaluation based on it. To preserve the privacy of service vendor $i$ regarding this information, we add Laplace noise with deviation $\lambda  = {{K\log (T)} \mathord{\left/
 {\vphantom {{K\log (T)} \varepsilon }} \right.
 \kern-\nulldelimiterspace} \varepsilon }$ to $f_{k^*,{x_i}(t)}^i(t)$ (line 7-9). Finally we update some counters and judge whether to partition the $C$ as described in \textbf{Phase 3} (line 11-14).

\begin{figure}[!htb]
 \vspace{-.5em}
\centering
\includegraphics[scale=.36]{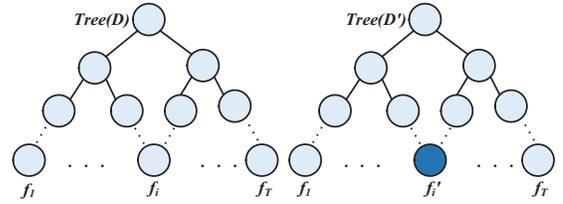}
\caption{An illustration of tree-based aggregation. $Tree(D)$ and $Tree(D')$ are two databases that differ in one component.}
\label{fig:digraph}
\vspace{-.5em}
\end{figure}
  \subsection{Algorithm Analysis}
 The properties of the proposed algorithm are analyzed in this subsection. For simplicity of presentation, we replace service vendors with learners. We prove that the regret is
 sublinear converged over the time, and our P-DAP guarantees differential privacy.

\subsubsection{Regret Bound}

For each subspace $C$, let ${\overline u _{k,C}} = {\sup _{x \in C}}{u_{k,x}}$ and ${\underline u _{k,C}} = {\inf _{x \in C}}{u_{k,x}}$. Let ${x^*}$ be the context at the center of the hypercube $C$. We define the optimal arm for subspce $C$ as ${k^*} = \mathop {\arg \max }\limits_{k \in {{\cal K}_i}} {u_{k,{x^*}}}$. Then the suboptimal arms for learner $i$ in subspace $C$ can be written as follows:
\begin{align}
{{\rm{S}}_{s,l,B}} = \left\{ {k:{{\underline u }_{k^*,C}} - {{\bar u}_{k,C}} > B{m^{ -
\alpha l}}} \right\},
\end{align}
where $B$ is a constant and $\alpha  > 0$. We will bound $B$ to get optimal solution. The regret in (2) can be written as the sum of three components:
\begin{align}
R\left( T \right) \le {R_o}\left( T \right) + {R_s}\left( T \right) + {R_n}\left( T \right),
\end{align}
where ${R_o}\left( T \right)$ is the regret due to selecting suboptimal arms from ${{\cal M}_i}$ by time $T$, ${R_s}\left( T \right)$ is the regret due to selecting
suboptimal arms from ${{\cal M}_{-i}}$  and ${R_n}\left( T \right)$ is the regret
of near optimal selections by time $T$. Next, we bound each of these terms separately.\vspace{1ex}

\textbf{Theorem 2.} \emph{For every level-l context subspace $C$, with control function ${G_1}\left( t \right) = {m^{2\alpha l}}ln (T)$, the expected regret due to
choosing suboptimal arm
$k \in {{\cal M}_i}$, will be bounded as follows:}
\begin{align}
\notag E\left[ {{\rm{Re}}g_{k,C}^o\left( T \right)} \right] &\le {m^{2\alpha l}}ln\left( T
\right) + \frac{{{\pi ^2}}}{3} \\&+ \frac{{2L{m^{ - \alpha l}}}}{\varepsilon }\left[ {ln\left( K
\right) + ln\left( T \right)} \right].
\end{align}
\begin{IEEEproof} The regret of $E[{\rm{Re}}g_{k,C}^o\left( T \right)]$ is due
to: 1) inherent gap of bandit algorithm between the optimal selections and the suboptimal
selections; 2) the  gap between approximately optimal reward applying exponential mechanism and suboptimal selections (line 11-14 in Algorithm 1):
\begin{align}
E\left[ {{\rm{Re}}g_{k,C}^o\left( T \right)} \right]\notag &\le E\sum\nolimits_{t = 1}^T {\left( {{u_{{k^*},x(t)}} - {u_{\varepsilon _u^\varepsilon \left( {x(t)} \right),x(t)}}} \right)} \\
\notag & \le E\sum\nolimits_{t = 1}^T {\left( {{u_{{k^*},x(t)}} - {u_{k,x(t)}}} \right)} \\
\notag &+ E\sum\nolimits_{t = 1}^T {\left( {{u_{k,x(t)}} - {u_{\varepsilon _u^\varepsilon \left( {x(t)} \right),x(t)}}} \right)} \\
 &= E\left[ {{\rm{Re}}g_{k,C}^1\left( T \right)} \right] + E\left[ {{\rm{Re}}g_{k,C}^2\left( T \right)} \right].
\end{align}

Next, we will bound the two part of the $E[{\rm{Re}}g_{k,C}^o\left( T \right)]$
separately:\vspace{1ex}

\textbf{Lemma 1.} \emph{ The inherent regret gap of bandit algorithm between optimal arms and suboptimal arms $E[{\rm{Re}}g_{k,c}^1\left( T \right)]$ is bounded as follows:}
\begin{align}
E\left[ {{\rm{Re}}g_{k,C}^1\left( T \right)} \right] \le {m^{2\alpha l}}ln\left( T \right) +
\frac{{{\pi ^2}}}{3}.
\end{align}

 \begin{IEEEproof}
 We denote ${F_k}(T)$ the
 number of times that suboptimal arm $k$ is selected by time $T$. For $x \in C$, let $\Delta {u_{k,C}} = {\overline u _{{k^*},C}} - {\underline u _{k,C}}$ be the gap of reward between suboptimal arm $k$ and optimal arm $k^*$ in subspace $C$. As initially defined, the regret of choosing suboptimal arm $k$ is the expected number of times when $k$ is selected times the gap of mean rewards. That is $E[ {{\rm{Re}}g_{k,C}^1\left( T \right)}] = \sum\nolimits_{t = 1}^T {{F_k}(T) \cdot \Delta {u_{k,C}}}  \le \sum\nolimits_{t = 1}^T {{F_k}(T)} $ for $\Delta {u_{k,C}} \le 1$. Inequality (11) results from the fact that ${F_k}(T)$  will not be larger than $ {m^{2\alpha l}}ln\left( T \right)$ with the high
 probability.  Now we discuss the result in inequality (11) under two
 circumstance.

\textbf{Case1.}  ${F_k}(T) \le {m^{2\alpha l}}ln (T)$. Under this circumstance, (11)
holds correctly. Now we focus on case2.\vspace{1ex}

\textbf{Case2.} ${F_k}({t_1}) = {m^{2\alpha l}}ln (T)$ when ${t_1} < T$. Then we have
\begin{align}
\begin{array}{l}
\notag {\rm{Re}}g_{k,C}^1\left( T \right) \le \sum\limits_{t = 1}^T {I\left( {k\;is\;picked\;at\;time\;t} \right)} \\
 \ \ \ \ \ \le {m^{2\alpha l}}ln\left( T \right) + \sum\nolimits_{t = {m^{2\alpha l}}}^T {I\left( {k\;is\;picked\;at\ time\ t} \right)}.
\end{array}
\end{align}

Next we will figure out the probability that $k$ is selected under Case2.

When $t > {m^{2\alpha l}}ln\left( T \right)$, if $k$ is selected, we have ${\overline r
_{k,C}}\left( t \right) \ge {\overline r _{k^*,C}}\left( t \right)$, this inequality holds
when at least one of the following holds:
\begin{align}
{\bar r_{k,C}}(t) \ge {\overline u _{k,C}} + {H_t},
\end{align}
\begin{align}
{\overline r _{{k^*},C}}(t) \le {\underline u _{{k^*},C}} - {H_t},
\end{align}
\begin{align}
 & \ \ \ \ \ \ \ \ \ \ \ \ \ \ \ \ \ \ \ {{\bar r}_{k,C}}(t) \ge {{\bar r}_{{k^*},C}}(t),\\& \ \ \ \ \ \ \ {{\bar r}_{k,C}}(t) < {{\bar u
\notag }_{k,C}} + {H_t},{{\bar r}_{{k^*},C}}(t){ > \underline u_{{k^*},C}} - {H_t}.
\end{align}

Then the probability when suboptimal arm $k$ is picked can be written as follows:
\begin{align}
\notag\ \ \ \ \ \ \ \ \mathbb{P} [&k\;is\;picked \cap Case2.] \\ \notag&\le \mathbb{P} [{{\bar r}_{k,C}}(t) \ge {{\bar
u}_{k,C}} + {H_t}] \\ \notag & + \mathbb{P}[{{\bar r}_{{k^*},C}}(t) \le {{}\underline u_{{k^*},C}} -
{H_t}] \\ \notag & + \mathbb{P} [{{\bar r}_{k,C}}(t) \ge {{\bar r}_{{k^*},C}}(t),\bar r{_{{k^*},C}}(t) <
{{\bar u}_{k,C}} + {H_t},\\&\ \ \ \ \ \ \ \ \ \ \  \ \ \ \ \ \ \  \ {{\bar r}_{{k^*},C}}(t) > \notag{{}\underline u_{{k^*},C}}
- {H_t}].
\end{align}

We denote $w_{k,C}^i({\rm{t}})$ the set of rewards of arm $k$ in subspace $C$. Let $O_{k,C}^i(t)$ be the event that at most $\frac{a}{n}$ samples in $w_{k,c}^i(t)$ are collected from
suboptimal process functions of the $k$-th arm. Different from
classical finite-time bandit theory, these samples are not identically distributed. Enlightened by \cite{cem}, in order
to facilitate our analysis of the regret, we also generate two different artificial i.i.d.
processes to bound the probabilities related to ${{\bar r}_{k,C}}(t)$, $k \in {{\cal M}_i}$.
The first one is the best process in which rewards are generated according to a bounded i.i.d.
process with expected reward ${{\bar u}_{k,C}}$, the other one is the
worst process in which the rewards are generated according to a bounded i.i.d. process with
expected reward ${\underline u _{k,C}}$.
Let $r_{k,C}^{best}(z)$ denote the sample mean of the $z$ samples from the best process and
$r_{k,C}^{worst}(z)$ denote the sample mean of the $z$ samples from the worst process. Thus, combining (7), for
any suboptimal arm we have: $\mathbb{P} ({{\bar r}_{k,C}}(t) \ge {{\bar r}_{{k^*},C}}(t),{{\bar
r}_{k,C}}(t) < {\overline u _{k,C}} + {H_t},{{\bar r}_{{k^*},C}}(t) > {\underline u
_{{k^*},C}}(t) - {H_t}) \le \mathbb{P} (\bar r_{k,C}^{best}(|w_{k,C}^i(t)|) \ge \bar
r_{k^*,C}^{worst}(|w_{k,C}^i(t)|) - \frac{a}{n}),\bar r_{k,C}^{best}(|w_{k,C}^i(t)|) <
{\overline u _{k,C}} + L{\left( {\frac{{\sqrt d }}{{{m^l}}}} \right)^\alpha} + {H_t} +
\frac{a}{n},\bar r_{{k^*},C}^{worst}(|w_{{k^*},C}^i(t)|) > {\underline u _{{k^*},C}} -
L{(\frac{{\sqrt d }}{{{m^l}}})^\alpha} - {H_t},case2)).$

Since $k$ is a suboptimal arm, we have  ${\underline u _{{k^*},C}} - {\bar u_{k,C}} > B{m^{ - \alpha l}},$ and :
\begin{small}
\[\bar r_{{k^{\rm{*}}},C}^{worst}(|w_{{k^{\rm{*}}},C}^i(t)|) > {\underline u _{{k^*},C}} - L{(\frac{{\sqrt d }}{{{m^l}}})^\alpha } - {H_t},\vspace{-1ex}\]
\[\bar r_{k,C}^{best}(|w_{k,C}^i(t)|) < {\bar u_{k,C}} + L{(\frac{{\sqrt d }}{{{m^l}}})^\alpha } + {H_t} + \frac{a}{n}.\vspace{-2ex}\]
\end{small}

Given the condition:
\begin{small}
\begin{equation}
2L{(\frac{{\sqrt d }}{{{m^l}}})^\alpha } + 2{H_t} + 2\frac{a}{n} - B{m^{ - \alpha l}} \le 0,
\end{equation}
\end{small}
we have :
\[\overline r _{k,C}^{best}(|w_{k,C}^i(t)|) < \overline r _{k^*,C}^{worst}(|w_{k,C}^i(t)|) -
\frac{a}{n},\]
which implies that suboptimal arms will hardly be selected by time:
\begin{small}
\begin{align}
\notag \mathbb{P} &[{\overline r _{k,C}}(t) \ge {\overline r _{{k^*},C}}(t),{\overline r _{k,C}}(t) <
{{\bar \mu }_{k,C}} + {H_t},{\overline r _{{k^*},C}}(t) > {{}\underline u_{{k^*},C}} - {H_t}] \\
 &= 0.
\end{align}
\end{small}
In Case2. we have $n \ge {m^{2\alpha l}}ln (t)$. In order to make (15) hold, we assign $B \ge 2L{(\frac{{\sqrt d }}{{{m^l}}})^\alpha } + 4$, $a = {m^{\alpha l}}ln (t)$, ${H_t} =
\frac{a}{n}$. Then, we have:
\[\mathbb{P} [{\bar r_{k,C}}(t) \ge {\bar \mu _{k,C}} + {H_t}] \le {e^{-2{{\left( {{H_t}} \right)}^2}{m^{\alpha l}}ln(T)}} = \frac{1}{{{t^2}}},\]
\[\mathbb{P} [{\bar r_{{k^*},C}}(t) \le {\underline u_{{k^*},C}} - {H_t}] \le {e^{-2{{\left( {{H_t}} \right)}^2}{m^{\alpha l}}ln(T)}} = \frac{1}{{{t^2}}}.\]
Thus, we have:

\[\begin{array}{*{20}{l}}
{\mathbb{P}(k\ is\ picked}&{\cap Case2.)\le \mathbb{P} ({{\bar r}_{k,C}}(t) \ge {{\bar u}_{k,C}} + {H_t})}\\
{}&{ + \mathbb{P} ({{\bar r}_{{k^*},C}}(t) \le {{}\underline u_{{k^*},C}}(t) - {H_t})}\\
{}&{ \le \frac{2}{{{t^2}}},}
\end{array}\]
then,
\[\begin{array}{*{20}{l}}
{E[{\rm{Re}}g_{k,c}^1(T)] \le {m^{2\alpha l}}ln (T) + \sum\nolimits_{{m^{2\alpha l}}ln (T)}^T {\frac{2}{{{t^2}}}} }\\
{ \le {m^{2\alpha l}}ln (T) + \frac{{{\pi ^2}}}{3}.}
\end{array}\]
\end{IEEEproof}
Before we derive Lemma 3, we provide a bound on the sensitivity of exponential mechanism.\vspace{1ex}

\textbf{Lemma 2.} \emph{The sensitivity of exponential mechanism is bounded is follows:}
\begin{equation}
\Delta u \le L{m^{ - \alpha l}}.
\end{equation}
\begin{IEEEproof} In our framework, ${x_1}$ and ${x_2}$ are two input data (users' context vectors),  which differ on at most one component. The unity function $u(x,k)$ represents the recommendation reward depending on input context $x$ and output video $k$. By Definition 3 and inequality (4), we have
\begin{align}
\Delta u \notag & = \mathop {\max }\limits_{k \in {{\cal M}_i}} \mathop {\max }\limits_{{x_1},{x_2}:{{\left\| {{x_1} - {x_2}} \right\|}_1} \le 1} \left| {q({x_1},k) - q({x_2},k)} \right| \\
\notag &= \mathop {\max }\limits_{k \in {{\cal M}_i}} \mathop {\max }\limits_{{x_1},{x_2}:{{\left\| {{x_1} - {x_2}} \right\|}_1} \le 1} \left| {{u_{k,{x_1}}} - {u_{k,{x_2}}}} \right| \\
\notag & \le \mathop {\max }\limits_{k \in {{\cal M}_i}} \mathop {\max }\limits_{{x_1},{x_2}:{{\left\| {{x_1} - {x_2}} \right\|}_1} \le 1} L{\left\| {x - x'} \right\|^\alpha } \le L{{\rm{m}}^{ - \alpha l}}.
\end{align}
\end{IEEEproof}

Combining Lemma 2 and Theorem 1, we can derive Lemma 3 as follows:\vspace{1ex}

\textbf{Lemma 3.}\emph{ The regret due to the near optimal reward when applying exponential
mechanism can be bounded as follows:}
\begin{equation}
E\left[ {{\rm{Re}}g_{k,C}^2\left( T \right)} \right] \le \frac{{2L{m^{ - \alpha
l}}}}{\varepsilon }\left[ {ln\left( K \right) + ln\left( T \right)} \right].
\end{equation}
\begin{IEEEproof}At each time slot, we do not choose the arm with highest reward. Instead, we
assign each arm a probability to be chosen. Thus, at each time slot, there exists the gap of
reward when applying the randomly selection. By using Theorem 1, in inequality (6), we have
$\left| R \right|{\rm{ = K}}$, $\left| {{R_{OPT}}} \right| = 1$ (we only have one optimal
arm). Then, we set  $t = ln (T)$.  Thus, at each time slot, we have the regret by randomly
selection as follows:
\[u(x,k) - u(x,\varepsilon _u^\varepsilon (x)) < \frac{{2\Delta u}}{\varepsilon }(ln(K) + ln (T)),\]
which holds with a probability less than $\frac{1}{T}$. Then, we have:
\begin{small}
\begin{align}
E\left[ {{\rm{Re}}g_{k,C}^2(T)} \right]\notag &{\rm{  = }}E\sum\nolimits_{t = 1}^T {\left( {{u_{k,x(t)}} - {u_{\varepsilon _u^\varepsilon \left( {x(t)} \right),x(t)}}} \right)} \\
 \notag &\le \sum\limits_1^T {\left[ {\Delta q} \right] \bullet \mathbb{P} [\Delta q < \frac{{2\Delta u}}{\varepsilon }(ln (K) + ln (T))} ] \\
 \notag &\le \sum\limits_1^T {\frac{{2\Delta u}}{\varepsilon }\left( {ln (K) + ln (T)} \right)}  \bullet \frac{1}{T}\\
 \notag &\le \frac{{2L{{\rm{m}}^{ - \alpha l}}}}{\varepsilon }\left( {ln (K) + ln (T)} \right),
\end{align}
\end{small}
where $\Delta q = u(x,k) - u(x,\varepsilon _u^\varepsilon (x)) = {u_{k,x(t)}} - {u_{\varepsilon _u^\varepsilon \left( {x(t)} \right),x(t)}} \le \frac{{2\Delta u}}{\varepsilon }(ln(K) + ln (T))$ denotes the regret bound of exponential mechanism selection at each time slot.\end{IEEEproof}

Combining Lemma 1, Lemma 2 and inequality (10), our Theorem 2 holds.\end{IEEEproof}
The above Theorem 2 implies that for $k \in {{\cal M}_i}$, the proposed algorithm make sure the suboptimal arms will be selected more than ${m^{2\alpha l}}\ln (T)$ with very small probability.
\vspace{1ex}

\textbf{Lemma 4.}\emph{ For $k \in {{\cal M}_{-i}}$, with control function ${G_2}(t) = {m^{2\alpha l}}\ln (t) + \frac{\Gamma }{4}{m^{\alpha l}}$ and ${G_1}\left( t \right) = {m^{2\alpha l}}ln (T)$,  we have the regret of choosing suboptimal $k$
in subspace C by time $T$ as follows:
\begin{small}
\begin{equation}
{\rm{E[Re}}g_{k,C}^{\rm{s}}(T)] \le {2m^{2\alpha l}}ln(T) + \frac{\Gamma }{4}{m^{\alpha l}} + \frac{{{\pi ^2}}}{3}(1 + \frac{K}{a})ln(T),
\end{equation}
\end{small}where $\Gamma $ is the near maximum value of the amount of total noise added by time $T$. We will bound $\Gamma $ in Lemma 5.}
\begin{IEEEproof} When we add Laplace noise to each time reward, our estimate of the actual
reward will be disturbed and our number of times that need to be played until finding the
optimal arm will be increased. But we demonstrate that, after each arm being trained ${G_1}({\rm{t}})$ times, there will be no more than
${m^{2\alpha l}}ln (T) + \frac{\Gamma }{4}{m^{\alpha l}}$ times to be tried before finding
the optimal arm with a high probability.

For $k \in {{\cal M}_{-i}}$, we define $k$ is the supoptimal arm, and ${k^*}$ is the optimal arm
for subspace $C$. At $t$-th time slot, suboptimal arm $k$ is selected over ${k^*}$ if ${\overline r _{k,C}}(t) \ge {\overline r _{{k^*},C}}(t)$ is true. Here, the reward ${\overline r _{k,C}}(t) \ge {\overline r _{{k^*},C}}(t)$ is the virtual
reward that include with noise for subspace $C$ of arm $k$. Thus, we denote ${\overline R
_{k,C}}(t)$ the true reward of arm $k$ for subspace $C$. Then suboptimal arm $k$ is selected, only if the following holds:
\begin{equation}
{\overline R_{k,C}}(t) + \frac{\Gamma }{{N_{k,C}^i(t)}} \ge {\overline R_{{k^*},C}}(t) + \frac{\Gamma }{{N_{{k^*},C}^i(t)}}.
\end{equation}
It can be easily shown that (17) is true, only if one of the following equations holds:
\begin{equation}
{\overline R_{k,C}}(t) \ge {\overline u_{k,C}} + {H_t},
\end{equation}
\begin{equation}
{\bar R_{{k^*},C}}(t) \le {\rm{ }}{\underline u _{{k^*},C}} - {H_t},
\end{equation}
\begin{align}
\notag &{{\overline R}_{k,C}}(t) < {\overline u _{k,C}} + {H_t},{{\overline R}_{{k^*},C}}(t) >
{\underline u _{{k^*},C}} - {H_t},\\
 &{{\overline R}_{k,C}}(t) + \frac{\Gamma }{{N_{k,C}^i(t)}} \ge {{\overline R}_{{k^*},C}}(t) +
\frac{\Gamma }{{N_{{k^*},C}^i(t)}}.
\end{align}

As we have discussed above for $k \in {{\cal M}_i}$, we also denote best process and worst process to
bound the probabilities. Then, we have£º
\begin{small}
\[\overline R_{k,C}^{best}(|w_{k,C}^i(t)|) < {\overline u _{k,C}} + L{(\frac{{\sqrt d
}}{{{m^l}}})^\alpha } + {H_t} + \frac{a}{n},\]\vspace{-1.5ex}
\[\overline R_{k,C}^{worst}(|w_{{k^*},C}^i(t)|) > {\underline u _{k^*,C}} - L{(\frac{{\sqrt d
}}{{{m^l}}})^\alpha } - {H_t}),\]\vspace{-2.2ex}
\[\overline R _{k,C}^{best}(|w_{k,C}^i(t)|) + \frac{\Gamma }{{N_{k,C}^i(t)}} \ge \overline R
_{k^*,C}^{worst}(|w_{{k^*},C}^i(t)|) + \frac{\Gamma }{{N_{{k^*},C}^i(t)}} - \frac{a}{n}.\]
\end{small}
When $k$ is a suboptimal arm, we have ${\underline u _{{k^*},C}} - {\bar u_{k,C}} > B{m^{ - \alpha l}}$.

Together imply that:
\begin{small}
\[2L{(\frac{{\sqrt d }}{{{m^l}}})^\alpha } + 2{H_t} + \frac{\Gamma }{{N_{k,C}^i(t)}} - \frac{\Gamma }{{N_{{k^*},C}^i(t)}} + 2\frac{a}{n} - B{m^{ - \alpha l}} \le 0.\]
\end{small}
For $n > N_{k,C}^i(t)$, ${H_t} = \frac{a}{n}$ and $B \ge
2L{(\frac{{\sqrt d }}{{{m^l}}})^\alpha } + 4$, then, we draw the conclusion that (20) holds when the following holds:
\begin{small}
\[2L{(\frac{{\sqrt d }}{{{m^l}}})^\alpha } + \frac{\Gamma }{{N_{k,C}^i(t)}} + 4\frac{a}{{N_{k,C}^i(t)}} - B{m^{ - \alpha l}} \le 0.\]
\end{small}

 Then we come to a conclusion that when $N_{k,C}^i({\rm{t)}} \ge {m^{2\alpha l}}ln (T) +
 \frac{\Gamma }{4}{m^{\alpha l}}$, the inequality (20) can not hold. (we use $S_C^i(t)$ denote
 this case), directly by the use of Chernoff bound, we can show that :
\begin{equation}
\mathbb{P} ({\overline R_{k,C}}(t) \ge {\overline \mu _{k,C}} + {H_t}) \le \frac{1}{{{t^2}}},
\end{equation}
\begin{equation}
\mathbb{P} ({\overline R_{{k^*},C}}(t) \le {\underline u_{{k^*},C}} - {H_t}) \le \frac{1}{{{t^2}}}.
\end{equation}
Let $O_{k,C}^i(t)$ be the event that at most $\frac{a}{n}$ samples in $w_{k,C}^i(t)$ are collected from
suboptimal process functions of the $k$-th arm. Obviously for any $k \in {{\cal M}_i}$, $O_{k,C}^i(t)
= \Omega $, while this is not always true for $k \in {{\cal M}_{-i}}$. Combining (17) and (18), for
$k \in {{\cal M}_{-i}}$, we have:
\[\mathbb{P}({O^i_{k.C}}(t),S_C^i(t)) \le \frac{2}{{{t^2}}}.\]

 For $k \in {{\cal M}_i}$ obviously we have  $\mathbb{P}(O_{k,C}^i{(t)^C} = 0)$ . For $k \in {{\cal M}_{-i}}$, let
 $Y_{k,C}^i(t)$ denote the random variable, the number of times suboptimal function $m$
 of for arm $k$ is chosen when event $S_C^i(t)$ holds. We have  $\{ O_{k,C}^i{(t)^C},S_C^i(t)\}  = \{ Y_{{\rm{k}},C}^i(t)
 \ge a\} $. Applying the Markov inequality, we have  $\mathbb{P}(O_{k,C}^i{(t)^C},S_C^i(t)) \le
 \frac{{E[Y_{k,C}^i(t)]}}{a}$. Let $E_{k,C}^i(t)$ be the event that a suboptimal processing
 function   $m \in {{\cal M}_k}$  is called by learner  $k$, when it is invoked by learner $i$ for
 the $t$-th time, we have
 \[Y_{k,C}^i(t) = \sum\nolimits_{t' = 1}^{w_{k,C}^i(t)} {I(E_{k,C}^i(t'))} ,\]
  and
 \[{\mathbb{P}\left[ {E_{k,C}^i\left( t \right)} \right] \le \sum\limits_{m \in {{\cal M}_{j}}} {\mathbb{P}\left( {{{\bar r}_{m,C}}\left( t \right) \ge \bar r_C^{*m}\left( t \right)} \right)} }\]
After each video $m \in {{\cal M}_k}$ has been fully explored by ${G_3}\left( t \right) = {m^{2\alpha l}}ln(t)/K$ times, as we have proofed in Lemma 1, we have
\[\mathbb{P}(E_{k,C}^i(t)) \le \sum\limits_{m \in {{\cal {M}}_k}} {2{e^{ - 2{{({H_t})}^2}{m^{2\alpha l}}ln(t)}}}  \le \frac{{2K}}{{{t^2}}}{e^K}.\]
Together imply that
\[E[Y_{k,C}^i(t)] \le \sum\limits_{t' = 1}^\infty  {\mathbb{P} (E_{k,C}^i(t'))}  \le
\sum\limits_{{\rm{t}} = 1}^\infty  {\frac{{2K}}{{{t^2}}}}{e^K} .\]
Therefore, from the Markov inequality we get
\[\mathbb{P} (O_{k,l}^i{(t)^C},S_C^i(t)) \le \frac{{E[Y_{k,C}^i(t)]}}{a} \le \frac{{{\pi ^2}}}{3} \cdot {e^K}\frac{K}{a}ln (T).\]

Then, for arm $k \in {{\cal M}_{-i}}$, 
we have
\begin{small}
\begin{align}
\notag E[&Reg_{k.C}^s(T)] \le \sum\limits_{t = 1}^T I (k\;is\;picked) \\
\notag &\le {2m^{2\alpha l}}ln(T) + \frac{\Gamma }{4}{m^{\alpha l}}ln(\Gamma ) + \sum\limits_{t
= 1}^T {\mathbb{P} } (S_C^i(t)) \\
\notag &\le {2m^{2\alpha l}}ln(T) + \frac{\Gamma }{4}{m^{\alpha l}}ln(\Gamma ) \\
\notag &+ \sum\limits_{t = 1}^T {\left[ {\mathbb{P}({O_{k.C}}(t),S_C^i(t)) + \mathbb{P}
(O_{k,C}^i{{(t)}^{\rm{c}}},S_C^i(t))} \right]}  \\
&\le {2m^{2\alpha l}}ln(T) + \frac{\Gamma }{4}{m^{\alpha l}}ln(\Gamma ) + \frac{{{\pi ^2}}}{3}{e^K}(1 +
\notag\frac{{K}}{a})ln(T).
\end{align}
\end{small}
\end{IEEEproof}\vspace{1ex}

\textbf{Lemma 5.}\emph{ For all arms $k \in {{\cal M}_{-i}}$ and all time step $t \in [T]$, $w.p. \ge 1 - \sigma $  (over the randomness), the amount of noise $\Gamma $ added in the total reward for $k$ till time $t$ is at most $\left| {{N_k}(t)} \right| \le \frac{{\theta {{\log }^2}(T)log\left( {{{\theta Tlog(T)} \mathord{\left/
 {\vphantom {{\theta Tlog(T)} \varphi }} \right.
 \kern-\nulldelimiterspace} \sigma }} \right)}}{\varepsilon }$, where $\theta $ is the number of arms belong to ${{\cal M}_{-i}}$}.
 \begin{IEEEproof}For the ease of notation, let ${R_k}(t)$ be the true total reward for arm $k$ until time $t$. As discussed above, ${N_k}(t) = {r_k}(t) - {R_k}(t)$ is a sum of at most $\log (T)$ Laplace distributed random variables $Lap(\frac{{\theta \log (T)}}{\varepsilon })$. By the tail property of Laplace distribution, we know that for a given random variable $x \sim Lap(\lambda )$, with probability $1 - \varphi $, $\left| x \right| \le \lambda \log (1/\varphi )$. So, with probability at least ${\left( {1 - {\varphi  \mathord{\left/
 {\vphantom {\varphi  {\log (T)}}} \right.
 \kern-\nulldelimiterspace} {\log (T)}}} \right)^{\log (T)}} \le 1 - \varphi $, $\left| {{N_k}(t)} \right| \le \frac{{\theta {{\log }^2}(T)log\left( {{{log(T)} \mathord{\left/
 {\vphantom {{log(T)} \varphi }} \right.
 \kern-\nulldelimiterspace} \varphi }} \right)}}{\varepsilon }$. Taking the union bound over all $k$-arms and all time step $T$ and setting $\varphi  = {\sigma  \mathord{\left/
 {\vphantom {\sigma  {(\theta T)}}} \right.
 \kern-\nulldelimiterspace} {(\theta T)}}$, we have $w.p. \ge 1 - \sigma $, for all $k \in {{\cal M}_{-i}}$ and for all $t \in [T]$, $\left| {{N_k}(t)} \right| \le \frac{{\theta {{\log }^2}(T)log\left( {{{\theta Tlog(T)} \mathord{\left/
 {\vphantom {{\theta Tlog(T)} \varphi }} \right.
 \kern-\nulldelimiterspace} \sigma }} \right)}}{\varepsilon }$.
 \end{IEEEproof}\vspace{1ex}

\textbf{Lemma 6.}\emph{ The regret due to choose near-optimal arms ${\mathop{\rm Re}\nolimits}
g_C^n(T)$ in each level-l subspace is bounded as follows:}
\begin{equation}
{\mathop{\rm Re}\nolimits} g_C^n(T) \le AB{m^{l(p - a)}}.
\end{equation}

\begin{IEEEproof}Due to the definition of near-optimal arms, regret due to selecting a
near-optimal arm is at
most $B{{\rm{m}}^{ - \alpha l}}$. Because there could be at most $A{{\rm{m}}^{pl}}$ slots for a level-$l$ subspace according to the partitioning rule, the regret
of this part is at most $AB{{\rm{m}}^{l(p - a)}}$.\end{IEEEproof}

Now, we combine the results in Lemma 4, Lemma 6 and Theorem 2 to obtain the complete regret
bound. The regret depends on the context arrival process and hence, we let \ $H_l^i(T)$ denote
the number
of level-$l$ subspaces that have been activated by time $T$ of learner $i$.\vspace{1ex}
 Before we derive Theorem 6, we provide a bound on the highest level of active subspace by time .\vspace{1ex}

\textbf{Theorem 3.}\emph{ The complete regret of our private distributed learning algorithm is
bounded by}
\begin{small}
\begin{align}
\begin{array}{*{20}{l}}
{R(T) \le \sum\nolimits_{k \in {{\cal M}_i}} {\sum\nolimits_l {H_l^i} (T) \cdot \left[ {{m^{2\alpha l}}ln(T) + \frac{{{\pi ^2}}}{3}} \right.} }\\
{\left. { \ \ \ \ \ \ \ \ \ \ \ \ \ \ \ \ \ \ \ \ \ \ \ \ \ \ \ \ \ \ \ \ \ \ \ \  + \frac{{2L{m^{ - \alpha l}}}}{\varepsilon }\left( {ln(K) + ln(T)} \right)} \right]}\\
{  \ \ \ \ \ \ \ \ \ \ \ \ \ \ + \sum\nolimits_{k \in {{\cal M}_{-i}}} {\sum\nolimits_l {H_l^i} } (T) \cdot \left[ {2{m^{2\alpha l}}ln(T) + \frac{\Gamma }{4}{m^{\alpha l}}} \right.}\\
{\left. {  \ \ \ \ \ \ \ \ \ \ \ \ \ \ \ \ \ \ \ \ \ \ \ \ \ \ \ \ \ \ \ \ \ \ \ \ \ \ \ \ \ \ \ \ \  + \frac{{{\pi ^2}}}{3}{e^K}\left( {1 + \frac{K}{a}ln(T)} \right)} \right]}\\
 { \ \ \ \ \ \ \ \ \ \ \ \ \ \ + \sum\nolimits_l {H_l^i} (T) \cdot AB{m^{l(p - a)}}.}
\end{array}
\end{align}
\end{small}\vspace{-2ex}
\begin{IEEEproof}
Combining the result of Lemma 1 and Lemma 3  it is easy to see that
${R_{\rm{o}}}(T)$ is bounded as follows:
\begin{small}
\[\begin{array}{*{20}{l}}
{{R_{\rm{o}}}(T) \le \sum\limits_l {H_l^i} (T)\sum\limits_{k \in {{\cal M}_i}}  \cdot  E[Reg_{k.C}^o(T)]}\\
{\ \ \ \ \ \ \ \ \  \le \sum\limits_l {H_l^i} (T) \cdot (M-1)\left[ {{m^{2\alpha l}}ln(T) + \frac{{{\pi ^2}}}{3}} \right.}\\
{\left. {\ \ \ \ \ \ \ \ \ \ \ \ \ \ \ \ \ \ \ \ \  + \frac{{2L{m^{ - \alpha l}}}}{\varepsilon }\left( {ln(K) + ln(T)} \right)} \right].}\\
{}
\end{array}\]
\end{small}

By applying Lemma 4, the ${R_s}(T)$ is bounded by
\begin{small}
\[\begin{array}{*{20}{l}}
{{R_s}(T) \le \sum\limits_{k \in {{\cal M}_{ - i}}} {\sum\limits_l {H_l^i} } (T) \cdot E[Reg_{k \cdot C}^{\rm{s}}(T)]}\\
{\ \ \ \ \ \ \ \ \  \le \sum\limits_{k \in {{\cal M}_{ - i}}} {\sum\limits_l {H_l^i} } (T) \cdot \left[ {{2m^{2\alpha l}}ln(T) + \frac{\Gamma }{4}{m^{\alpha l}}} \right.}\\
{\left. { \ \ \ \ \ \ \ \ \ \ \ \ \ \ \ \ \ \ \ \ \ \ \ \ \ \ \ \ \ \ \ \ \ + \frac{{{\pi ^2}}}{3}{e^K}\left( {1{\rm{ + }}\frac{K}{a}ln (T)} \right)} \right].}
\end{array}\]
\end{small}

Finally, ${R_n}(T)$ is bounded by
\begin{small}
\[{R_n}(T) \le \sum\limits_l {H_l^i} (T) \cdot E[{\rm{Re}}g_C^n(T)] \le \sum\limits_l {H_l^i} (T) \cdot AB{m^{l(p - a)}}\]
\end{small}
Theorem is resulted of the summing of above three equation.
\end{IEEEproof}

The following corollary establishes the regret bound when the context arrivals are uniformly
distributed over the entire context space. This is the worst-case scenario because the
algorithm has to learn over the entire context space. Before we derive Corollary 2, we provide a bound on the highest level of active subspace by time.\vspace{1ex}

\textbf{Lemma 7.}\emph{  Given a time T , the highest level of active subspace
is at most $\left\lceil {{{\log }_m}(\frac{T}{A})/P} \right\rceil  + 1$.}\vspace{1ex}

\begin{IEEEproof}It is easy to see that the highest possible level of
active subspace is achieved when all requests by time have the
same context. This requires $A{{\rm{m}}^{{l_{\max }}}} \le T$. Therefore, ${l_{\max }} =
\left\lceil {{{\log }_m}(\frac{T}{A})/P} \right\rceil  + 1$.\end{IEEEproof}\vspace{2ex}

\textbf{Corollary 2.} \emph{If the context arrival by time T is uniformly distributed over the context space, and we set the partition parameter $p$ much larger than similarity parameter $\alpha$ we have:}
\begin{small}
\begin{align}
\begin{array}{l}
R\left( T \right) \le {R_o}\left( T \right) + {R_s}\left( T \right) + {R_n}\left( T \right)\\
 \ \ \ \ \ \ \ \ \le {(\frac{T}{A})^{\frac{{d + 2\alpha }}{{d + p}}}} \cdot {m^{d + 2\alpha }}ln(T)\left( {K + M - 1} \right)\\
  \ \ \ \ \ \ \ \ + {(\frac{T}{A})^{\frac{{d + \alpha }}{{d + p}}}} \cdot {m^{d + \alpha }}AB\\
  \ \ \ \ \ \ \ \ + {(\frac{T}{A})^{\frac{{d + p - \alpha }}{{d + p}}}} \cdot {m^{d + p - \alpha }}\frac{\Gamma }{4}\\
 \ \ \ \ \ \ \ \  + {(\frac{T}{A})^{\frac{d}{{d + p}}}} \cdot {m^d} \cdot \frac{{{\pi ^2}}}{3}(K + (M - 1){e^K})\\
 \ \ \ \ \ \ \ \  + {(\frac{T}{4})^{\frac{{d - \alpha }}{{d + p}}}}{m^{d - \alpha }}[\frac{{2KL}}{\varepsilon }(ln(K) + ln(T)) + \frac{{{\pi ^2}}}{3}(M-1){e^K}ln(T)].
\end{array}
\end{align}
\end{small}
\begin{IEEEproof}First we calculate the highest level of subspace when context arrivals are uniform. In
the worst case, all level $l$ subspaces will stay active, and then they are deactivated until
all level-$(l + 1)$ subspaces become active and so on. Let ${l_{\max }}$ be the maximum level
subspace under this scenario. Because there must be some time $T' < T$ when all subspaces are
level subspaces, we have
\[{m^{dl}}A{m^{pl}} < T,\]
where ${m^{dl}}$ is the maximum number of level $l$ subspaces and $A{m^{pl}}$
is the maximum number of time slots that belong to a level $l$
subspace. Thus, we have ${l_{\max }} < \frac{{{{\log }_m}(\frac{T}{A})}}{{d + p}} + 1$ .
Combining this conclusion with the regret bound in Theorem 3, we get Corollary 2.
\end{IEEEproof}

We have shown that the regret upper bound  of our private distributed learning model is sublinear in
time, implying our computing service vendors can select optimal videos by time.  Also, fast convergence to optimal is favorable to dynamically changing big data environments.\vspace{1ex}

\subsubsection{Differential Privacy}
We finally prove that our algorithm can preserve privacy of user's contextual information and the that of each service vendor's videos.\vspace{1ex}

\textbf{Theorem 4.}\emph{ The Algorithm 1 can preserves $(\varepsilon ,0)$-differential privacy for user's contextual information.}\vspace{1ex}
\begin{IEEEproof}Let ${x_1}$ and ${x_2}$ be two input context  vectors that differ in one
single attribute, $\mu $ denote the reward of exponential mechanism, $R$ denotes the output (sequence of selected videos)
space of exponential mechanism. Then $R = \{
{k_{1,}}{k_2},...,{k_{M + K - 1}}\} $. We suppose that the same user's data stream has come
for $N$ times over time arbitrary sequence ${\rm{\{
}}{{\rm{t}}_1}{\rm{,}}{{\rm{t}}_2}{\rm{,}}...{\rm{,}}{{\rm{t}}_N}{\rm{\} }}$, as a result, our algorithm selected an arbitrary sequence of arms such that
${{M_E}({x_1},\mu ,R) = \{ {k_1},{k_2},...,{k_N}\} }$ at the time sequence. We denote $\mu
({x_1},{k_i})$ the mean reward of arm ${k_i}$ for context ${x_1}$ at time ${t_i}$. In
our algorithm $\mu ({x_1},{k_i})$  equals ${\overline r _{{k_i},C}}({t_i})$. $C$ is the active
subspace to which the context ${x_1}$ belongs at time. If ${x_1}$ and ${x_2}$ belong to the
same subspace $C$ at time ${t_i}$, then $\mu ({x_1},{k_i}) = \mu ({x_2},{k_i})$. We construct a
function $I({t_1},{x_1},{x_2})$. When ${x_1}$, ${x_2}$ belong to the same active subspace,
the value of the function equals one, otherwise zero.
 We consider the relative probability of our algorithm for given context ${x_1}$ and ${x_2}$:

 \begin{small}
\[\begin{array}{l}
\frac{{{\mathbb{P}}[{M_E}({x_1},\mu ,R) = \{ {k_1},{k_2},...,{k_N}\} ]}}{{{\mathbb{P}}[{M_E}({x_2},\mu ,R) = \{ {k_1},{k_2},...,{k_N}\} ]}}\\
 = \prod\limits_{i = 1}^N {{{( {\frac{{exp(\frac{{\varepsilon '\mu ({x_1},{k_i})}}{{2\Delta \mu }})}}{{\sum\limits_{k' \in R} e xp(\frac{{\varepsilon '\mu ({x_1},k')}}{{2\Delta \mu }})}}} )} \mathord{\left/
 {\vphantom {{\left( {\frac{{exp(\frac{{\varepsilon '\mu ({x_1},{k_i})}}{{2\Delta \mu }})}}{{\sum\limits_{k' \in R} e xp(\frac{{\varepsilon '\mu ({x_1},k')}}{{2\Delta \mu }})}}} \right)} {\left( {\frac{{exp(\frac{{\varepsilon '\mu ({x_2},{k_i})}}{{2\Delta \mu }})}}{{\sum\limits_{k' \in R} e xp(\frac{{\varepsilon '\mu ({x_2},k')}}{{2\Delta \mu }})}}} \right)}}} \right.
 \kern-\nulldelimiterspace} {( {\frac{{exp(\frac{{\varepsilon '\mu ({x_2},{k_i})}}{{2\Delta \mu }})}}{{\sum\limits_{k' \in R} e xp(\frac{{\varepsilon '\mu ({x_2},k')}}{{2\Delta \mu }})}}} )}}} \\
 = \prod\limits_{i = 1}^N e xp(\frac{{\varepsilon '(\mu ({x_1},{k_i}) - \mu ({x_2},{k_i}))}}{{2\Delta \mu }}) \cdot (\frac{{\sum\limits_{k' \in R} e xp(\frac{{\varepsilon '\mu ({x_2},k')}}{{2\Delta \mu }})}}{{\sum\limits_{k' \in R} e xp(\frac{{\varepsilon '\mu ({x_1},k')}}{{2\Delta \mu }})}})\\
 = \prod\limits_{i = 1}^N e xp(\frac{{\varepsilon '(\mu ({x_1},{k_i}) - \mu ({x_2},{k_i}))}}{{2\Delta \mu }}) \cdot (\frac{{\sum\limits_{k' \in R} e xp(\frac{{\varepsilon '\mu ({x_2},k')}}{{2\Delta \mu }})}}{{\sum\limits_{k' \in R} e xp(\frac{{\varepsilon '\mu ({x_1},k')}}{{2\Delta \mu }})}})\\
 \le \prod\limits_{i = 1}^N e xp\left( {\frac{{\varepsilon '}}{2} \cdot I({t_i},{x_1},{x_2})} \right) \cdot exp\left( {\frac{{\varepsilon '}}{2} \cdot I({t_i},{x_1},{x_2})} \right)\\
 \ \ \ \ \ \times ( {\frac{{\sum\limits_{k' \in R} e xp(\frac{{\varepsilon \mu ({x_1},k')}}{{2\Delta \mu }})}}{{\sum\limits_{k' \in R} e xp(\frac{{\varepsilon \mu ({x_1},k')}}{{2\Delta \mu }})}}} )\\
 = \prod\limits_{i = 1}^N e xp(\varepsilon ' \cdot I({t_{\rm{i}}},{x_1},{x_2}))\\
 \le exp(N\varepsilon ')\\
 = exp(\varepsilon ).
\end{array}\]
\end{small}

Thus, the theorem follows.\end{IEEEproof}\vspace{1ex}

\textbf{Theorem 5.} \emph{ The Algorithm 2 can preserve  $(\varepsilon ,0)$-differential privacy for service vendors' videos}.

\begin{IEEEproof} For $k \in {{\cal M}_{-i}}$ and subspace $C$, let $\left[ {T'} \right] = \{ 1,...,T'\} $ denotes the sequence
of time slots that videos is selected for simplicity, where $T' < T$.  let $D = \left\langle
{{f_1},...,{f_T}} \right\rangle $ be a data set of true rewards. We call a data set $D'$
neighbor of $D$ if it differs from $D$ in exactly one reward. We define ${F_t}(C)$ the virtual outcome (reward with noise added), then we have, at each round, the probability of same outcome for different arm ${k_1}$ and ${k_2}$:
\[\begin{array}{l}
\frac{{\mathbb{P} [{M_L}({k_1},t) = {F_t}(C)]}}{{\mathbb{P} [{M_L}({k_2},t) = {F_t}(C)]}} = \frac{{\exp ( - \frac{{\varepsilon '\left| {{f_t}({k_1}) - {F_t}(C)} \right|}}{{\Delta f}})}}{{\exp ( - \frac{{\varepsilon '\left| {{f_t}({k_2}) - {F_t}(C)} \right|}}{{\Delta f}})}}\\
 = \exp \left( {\frac{{\varepsilon '}}{{\Delta f}}(\left| {{f_t}({k_2}) - {F_t}(C)} \right| - \left| {{f_t}({k_1}) - {F_t}(C)} \right|)} \right)\\
 \le exp(\frac{{\varepsilon '}}{{\Delta f}}\left| {{f_t}({k_1}) - {f_t}({k_2})} \right|)\\
 = exp(\frac{{\varepsilon '}}{{\Delta f}}{\left\| {{f_t}({k_1}) - {f_t}({k_2})} \right\|_1})\\
 \le exp(\varepsilon ').
\end{array}\]

In our problem model, the proposed algorithm only accesses the reward for its computation via the tree based aggregation scenario. Learner $i$ maintains $M - 1$ trees for other learner's reward sets respectively. Each tree guarantee  $\varepsilon ' = {\varepsilon  \mathord{\left/{\vphantom {\varepsilon  {(M - 1)}}} \right.\kern-\nulldelimiterspace} {(M - 1)}}$ differential privacy. With the composition property stated in Corollary 1, we can draw the conclusion that our algorithm 2 is $\varepsilon $-differential private.
 \end{IEEEproof}
 Theorem 4 shows that the attributes (e.g., social status, hobby and age) in users' sensitive context vectors cannot be inferred from the recommended results. The proof of Theorem 5 supports that the service vendors fail to extract information about videos in neighbor service vendors' repositories by the rewards.
In summary, our Theorem 4 and Theorem 5 prove that the proposed algorithm P-DAP can preserve the both privacy of users and service vendors synchronously.

\section{Geometric Differential Privacy}
In the previous section, we preserve privacy to the same extend for all subspaces. That is to say, we set the same value of $\varepsilon $ for the whole context space. This section presents our refined \emph{geometric differentially private} model. Considering the sparsity and heterogeneity of big data, some context subspaces are scattered with countless data points, however, other subspaces are nearly blank. A large and increasing number of statistical analyses can be done in a differential private manner while adding little noise. As also declared in \cite{DP}, ``the larger the dataset, the less a given amount of blurring will affect utility''.  Thus, our geometric differential private algorithm varies the amount of noised add to subspaces according to the size of each subspace. To be specific, we decrease the privacy level (larger value of $\varepsilon $) when the density of datasets increased (l denotes the density
of subspaces). In this way, the performance loss due to the randomness brought by differential privacy can be reduced extensively. For current active subspaces, we set different value of $\varepsilon $ related to the density $l$ of them. Specifically, we increase the value of $\varepsilon $ when $l$ increases. Fig. 5 gives an illustration of this method. For simplicity, we take the one-dimensional context space for instance. Leaf nodes presented in Fig. 5 are current active subspaces, we set different value of $\varepsilon $ related to the density $l$ of each subspaces.
\begin{figure}[!htb]
 \vspace{-.5em}
\centering
\includegraphics[scale=.40]{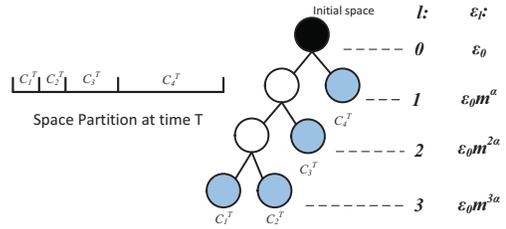}
\caption{An illustrative example of geometric private model: For simplicity, we assume dimension of context space $d = 1$. The left segment shows the partition pattern. The right tree structure shows the partition process, where blue leaf nodes denote the active subspaces. Subspaces with different level-$l$ get different value of $\varepsilon $.}
\label{fig:digraph}
\vspace{-.5em}
\end{figure}

The modified method works as follows. After we get enough context samples, we already have accurate estimations for rewards. From now on, for each context arrival, we first figure out to which subspace it belongs. Then we judge the level $l$ of the subspace and set $\varepsilon  = {\varepsilon _0}{m^{\alpha l}}$ for level-$l$ subspaces, where $m$ and $\alpha $ are constants as we have defined previously.

\textbf{Theorem 6.} \emph{Geometric differential privacy has a lower regret bound than uniform differential privacy as follows:}
\begin{equation}
\begin{array}{l}
{R^G}(T) \le R(T) - \left( {{{(\frac{T}{A})}^\alpha }{m^\alpha } - 1} \right)\left( {{A_1}{{(\frac{T}{A})}^{\frac{{d - 2\alpha }}{{d + p}}}}{m^{d - 2\alpha }}} \right.\\
\left. { \ \ \ \ \ \ \ \ \ \ \ \ + {A_2}{{(\frac{T}{A})}^{\frac{{d - \alpha }}{{d + p}}}}{m^{d - \alpha }}} \right),
\end{array}
\end{equation}
where ${{A_1}}$ and ${{A_2}}$ are two constants. When time $T$ goes into infinity, the value of the second term on the right side of the inequality will increase exponentially. Thus, the result of Theorem 6 proves that our geometric differential privacy has greatly reduced the regret bound.
\begin{IEEEproof}
We set $\varepsilon  = {\varepsilon _0}{m^{\alpha l}}$ and the amount of noise $\Gamma  = \frac{{{\Gamma _0}}}{{{\varepsilon _0}{m^{\alpha l}}}}$ in the geometric differential privacy method. Thus, we have:
\[\begin{array}{*{20}{l}}
{{R^G}(T) \le R(T) - \sum\nolimits_{l = 1}^{{l_{\max }}} {(M - 1)} {m^{\alpha l}}(\frac{{{\Gamma _0}}}{{{\varepsilon _0}}} - \frac{{{\Gamma _0}}}{{{\varepsilon _0}{m^{\alpha l}}}})}\\
{ - \sum\nolimits_{l = 1}^{{l_{\max }}} {K\left( {ln(T) + ln(T)} \right)}  \cdot \left( {\frac{{2L{m^{ - \alpha l}}}}{{{\varepsilon _0}}} - \frac{{2L{m^{ - \alpha l}}}}{{{\varepsilon _0}{m^{\alpha l}}}}} \right)}\\
{ \le \left( {{{(\frac{T}{A})}^\alpha }{m^\alpha } - 1} \right)\left( {\frac{{K\left( {ln(T) + ln(T)} \right)}}{{{\varepsilon _0}}}{{(\frac{T}{A})}^{\frac{{d - 2\alpha }}{{d + p}}}}{m^{d - 2\alpha }}} \right.}\\
{\left. { + \frac{{{\Gamma _0}}}{{{\varepsilon _0}}}{{(\frac{T}{A})}^{\frac{{d - \alpha }}{{d + p}}}}{m^{d - \alpha }}} \right).}
\end{array}\]

For simplicity, we use ${A_1}$ and ${A_2}$ denote ${\frac{{K\left( {ln (T) + ln (T)} \right)}}{{{\varepsilon _0}}}}$ and ${\frac{{{\Gamma _0}}}{{{\varepsilon _0}}}}$ respectively. Here the Theorem 6 holds.
\end{IEEEproof}

\section{Experimental Results and Analysis}
In this section, we demonstrate the theoretical regret bounds for our algorithms with empirical results based on very large real-world datasets, which includes massive multimedia data and social media users-generated big data.  We show that: 1) regret bounds are sublinear converged over time; 2) Our differentially private methods work well and do not come at the expense of recommendation accuracy; 3) \emph{Geometric differentially private} method has a lower regret bound and higher accuracy. Finally, we use users' context vectors refined from real datasets  to test the recommendation accuracy of our algorithms.
\subsection{Experimental Setup}
To evaluate the performance of our recommendation system, training data and test data about users and videos should be gathered. We collect numerous user context vectors extracted from large real datasets in Sina Microblog, a popular online social networking site in China. This datasets contain users' social profiles and multimedia content they shared. We also extract public information from Youku, a prevalent video sharing site (VSS) in China, such as video attributes, popular videos. After preprocessing, around 74000 video items, 578000 user context vectors with 13900-dimension are stored.

For simplicity, we deploy the recommendation system on a small-sized framework with four distributed video service vendors. Using collected video data, we constructed a set of 1000 videos for each service vendor respectively, Following the real situation, we arrange different video items for different service vendors. We randomly sample 200000 users ( context vectors) from our stored datasets, and input these vectors to our simulative recommendation system sequentially. When receiving user arrival, service vendor selects a particular video to recommend. At the end of this time slot, the reward of this selection, a binary random number (equal 0 or 1), is produce, to imitate the result of user's click action. Since our scheme appertains to the class of online distributed learning techniques, we will compare our scheme against several previous approaches:
\begin{itemize}
\item Centralized learning with adaptive partition (CAP)\cite{JX}: There is only one learner in this centralized framework who partitions the context space dynamically over time according to the number of user arrivals.
\item Distributed learning with uniform partition (DUP)\cite{cem}: This distributed framework contains multiple cooperative learners. But all of them uniformly partition context space initially. No partition process is involved over time.
\item Distributed learning with adaptive partition (DAP): This is the primal model of the proposed P-DAP. Multiple learners in this distributed framework adaptively partition the context space over time (No privacy preservation in this model).
\end{itemize}

Finally, to thoroughly analyze the performance of our proposed algorithms, we logically deploy our experiment by the following 4 steps:

\textbf{Step 1.} We first compare our primal model DAP with previous work, i.e., CAP \cite{JX} and DUP\cite{cem}. We input sampled 200000 users' context vectors sequentially into these three models respectively. That is to say, each model will receive same input datasets with 200000 elements. We plot the regrets and the average regrets (to evaluate the convergence rate) of each model. Afterwards, we extracted four groups (with different size) of user context vectors from collected real datasets. Then, we input these four groups context vectors into CAP, DUP and DAP to test the performance of each model.

\textbf{Step 2.} We construct our differentially private model (P-DAP) based on step 1. As for each vendors' own user, arms (videos and other service vendors) are randomly selected according to computed probabilities. Simultaneously, Laplace noise is added when recommending videos to other service vendors' users. To prove the smooth trade-off between privacy and accuracy in our P-DAP, we vary the privacy constant $\varepsilon $ from 0.01 to 1 and compare them with non-private model (DAP). Finally, we use our extracted four groups of context vectors to test the accuracy of these models.

\textbf{Step 3.} To prove the lower regret of geometric differential private method  (GP-DAP), we set different value of $\varepsilon $ for different context subspaces. To be specific, the value of $\varepsilon $ wax with the decrease of the density of data points in each subspace. Then, we compare the regrets of GP-DAP and P-DAP ($\varepsilon {\rm{ = }}0.01$) over time.
 \begin{figure}
\begin{minipage}{0.5\linewidth}
\centering
\includegraphics[width=1.9in]{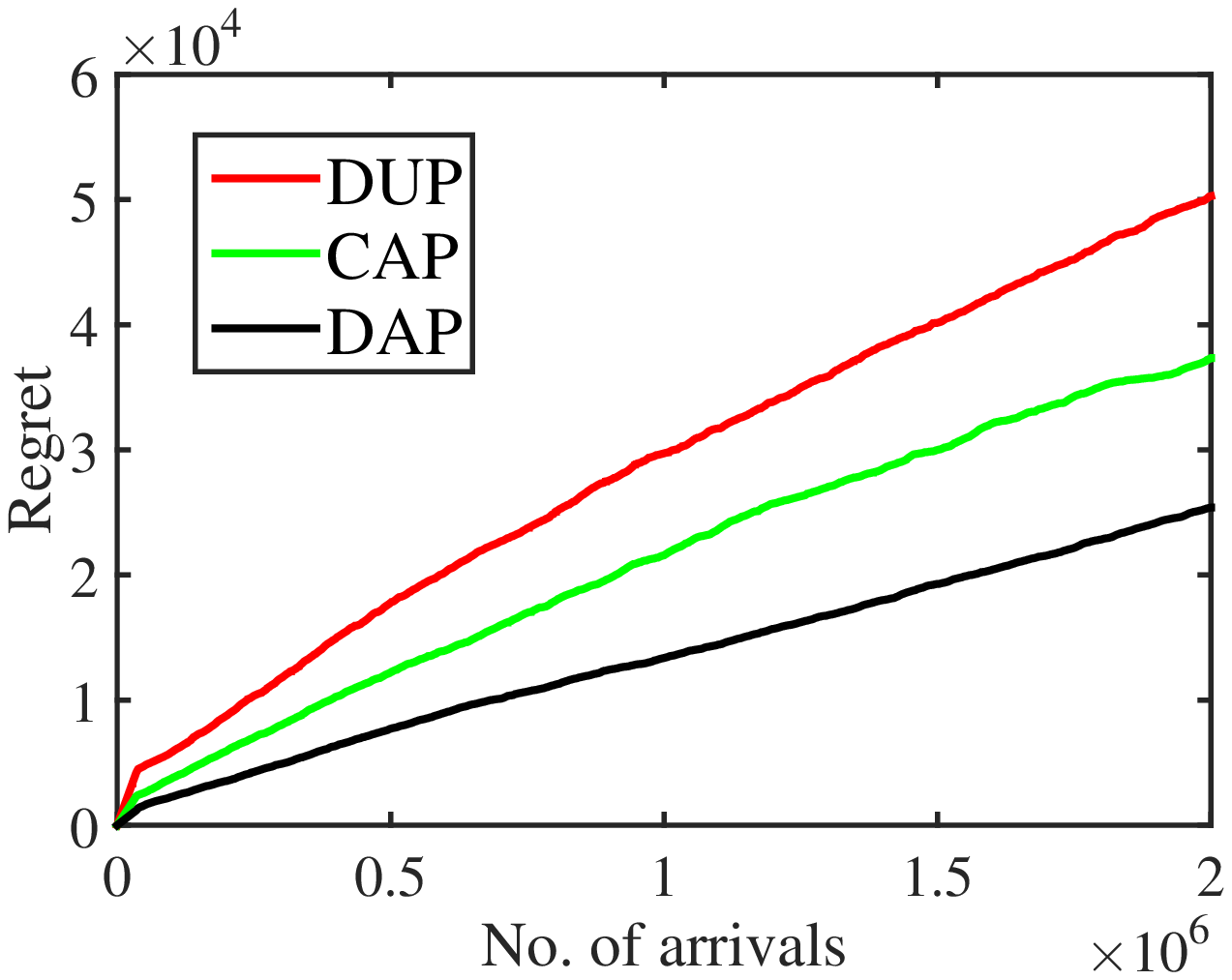}
\centering{(a) Regrets }
\label{fig:side:a}
\end{minipage}%
\begin{minipage}{0.5\linewidth}
\centering
\includegraphics[width=1.9in]{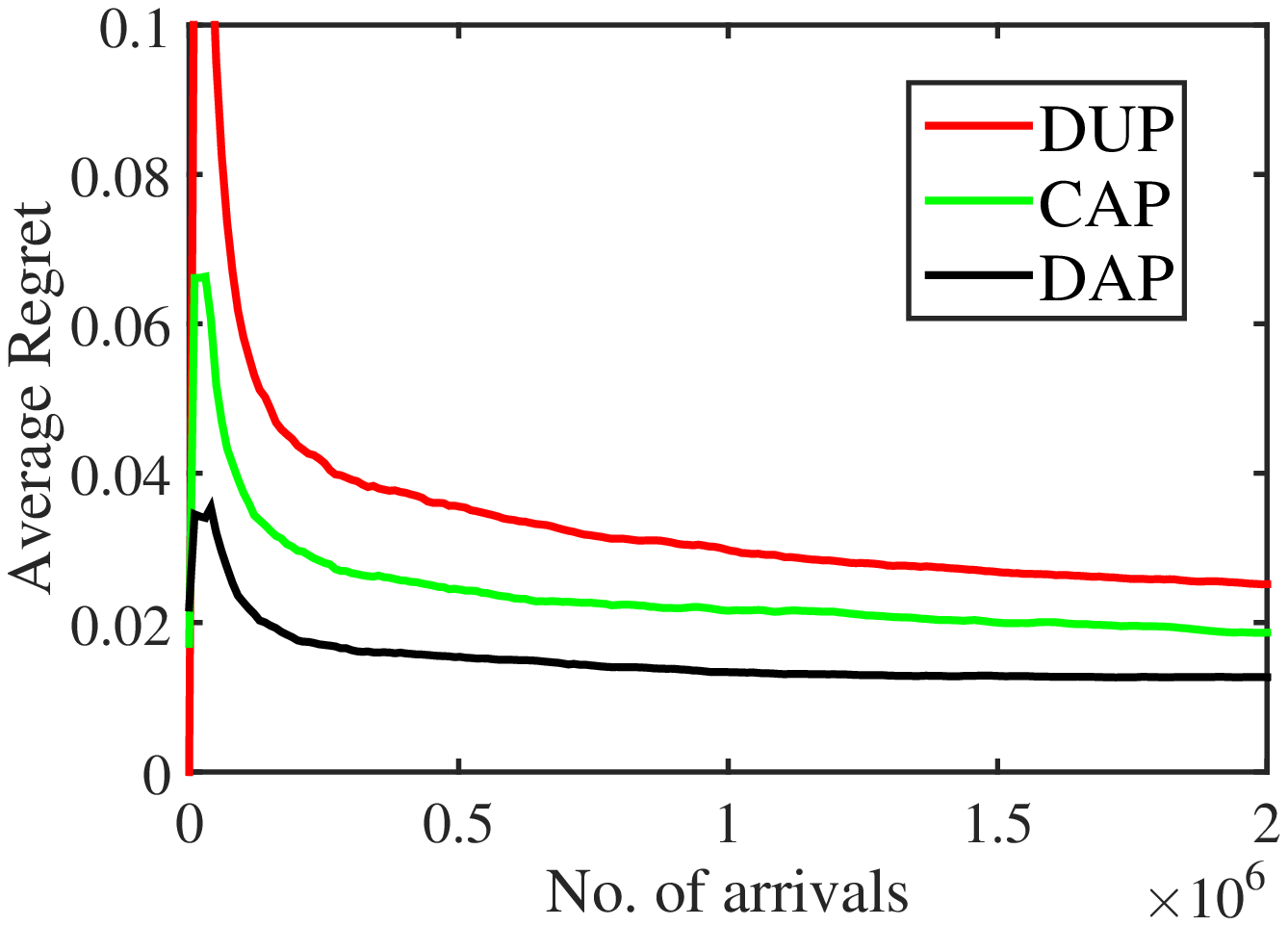}
\centering{(b) Average Regrets}
\label{fig:side:b}
\end{minipage}
\caption{Regrets in CAP, DUP and DAP}
\end{figure}

 \begin{table}
 \vspace{-.5em}
\centering
\begin{tabular}{|C{1.625cm}|C{1.625cm}|C{1.625cm}|C{1.625cm}|}
\hline
N & CAP & DUP & DAP \\
\hline
5000 & 70.32\% & 73.32\% & 87.34\% \\
\hline
10000 & 75.36\% & 74.52\% & 90.21\% \\
\hline
20000 & 76.88\%	& 74.78\% & 91.02\% \\
\hline
50000 & 77.34\%	& 75.08\% & 92.17\% \\
\hline
\end{tabular}
\caption{Average accuracies of DAP, CAP and DUP }
\end{table}

\subsection{Results and Analysis}

 We first evaluate DAP's performances in terms of regret loss and average regret loss in Step 1. In the meanwhile, we compare DAP, CAP and UAP and plot the regret lines in Fig. 6.

 Fig. 6 (a) shows the comparison with DAP, CAP and DUP in terms of regrets, where the horizontal axis is the number of user arrivals. From the tendency of ``Regret'' lines, we can draw the conclusion that the regret of DAP is sublinear converged over time. And obviously, DAP has lower regret loss than DUP and CAP all the time. Fig. 6 (b) records the average regrets (normalized by number of arrivals) of DAP, CAP and DUP, where the horizontal axis is the number of user arrivals. As we can see, our primal model DAP converges fast and has lower average accuracy then CAP and DUP. Also, results show the average regret of DAP in the tail of lines is extremely small (smaller than 0.02 per user).

 Table II records the average accuracies (total reward divided by number of arrivals) in our tested process, where $N$ represents the number of context vectors used by test. We find that as the number of arrivals increased, the average accuracies of each model get promoted as well. This could be resulted from the fact that systems trained better as number of samples increased. Also, we can read from the table that the average accuracy of our DAP can reach up to 92${\rm{\% }}$, but neither those of CAP nor DUP can exceed 80 ${\rm{\% }}$. Finally, We can draw the conclusion that DAP outperforms CAP and DUP.

\begin{figure}
\begin{minipage}{0.5\linewidth}
\centering
\includegraphics[width=1.9in]{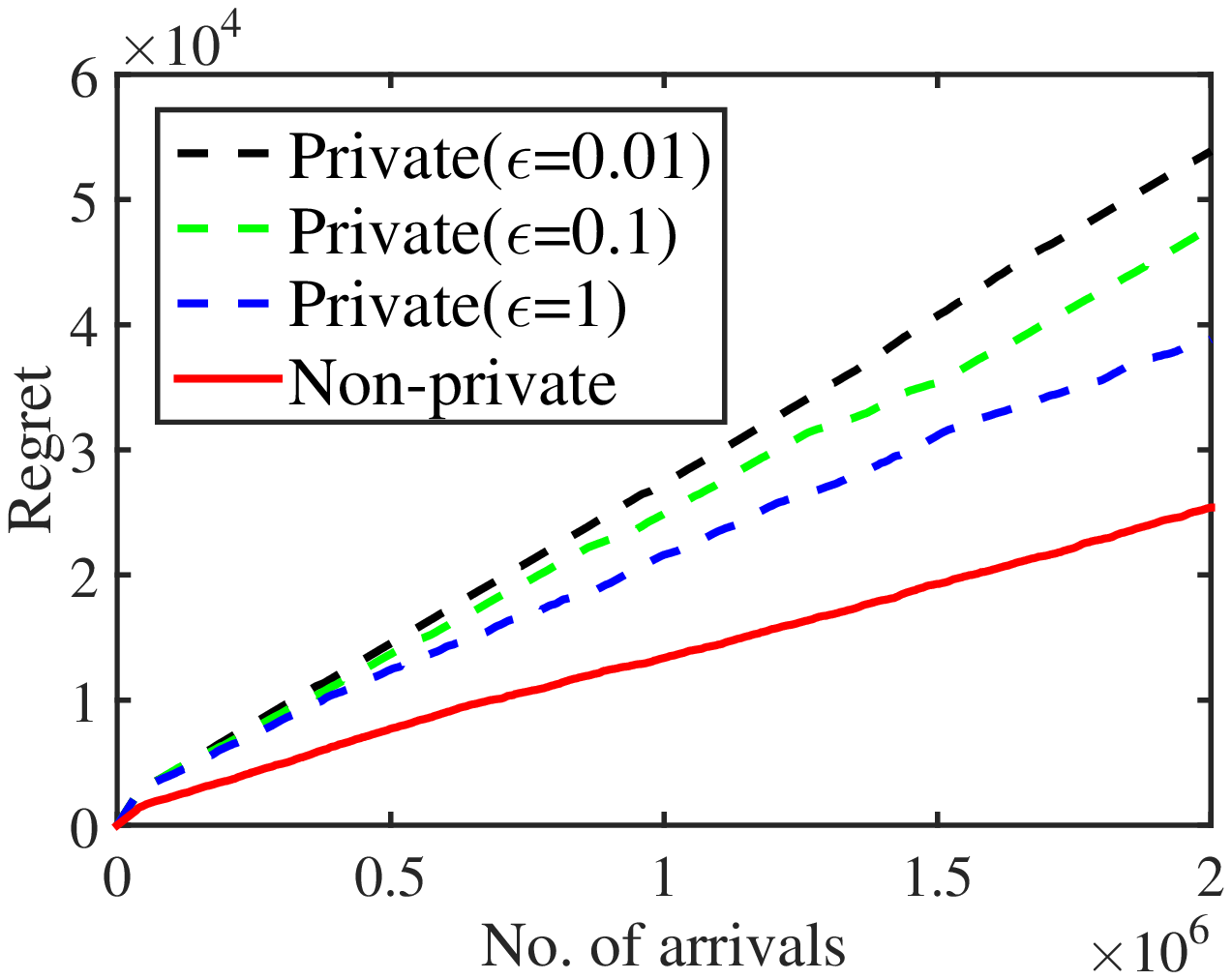}
\centering{(a) Regret}
\label{fig:side:a}
\end{minipage}%
\begin{minipage}{0.5\linewidth}
\centering
\includegraphics[width=1.9in]{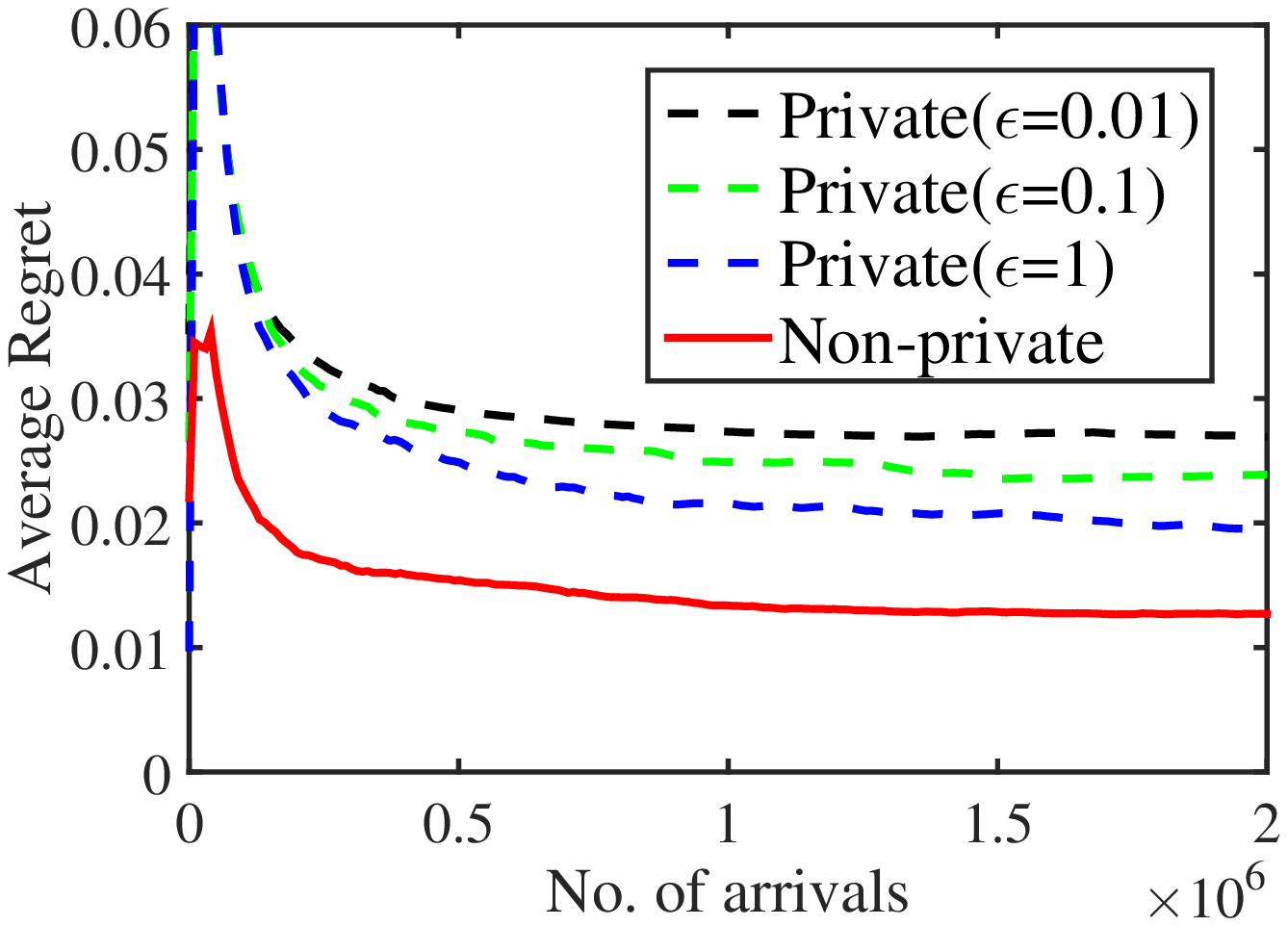}
\centering{(b) Average Regret}
\label{fig:side:b}
\end{minipage}
\caption{Regrets in P-DAP and DAP}
\end{figure}
\begin{table}
 \vspace{-.5em}
\centering
\renewcommand{\multirowsetup}{\centering}
\begin{tabular}{|C{1.3cm}|C{1.3cm}|C{1.3cm}|C{1.3cm}|C{1.3cm}|}
\hline
  & \multicolumn{3}{|c|}{P-DAP} & \multirow{2}{*}{DAP} \\
\cline{1-4}
N & $\epsilon$=0.01 & $\epsilon$=0.1 & $\epsilon$=1 & \\
\hline
5000 & 77.54\% & 77.78\% & 78.96\% & 86.84\% \\
\hline
10000 & 78.75\% & 79.35\% & 79.24\% & 89.28\% \\
\hline
20000 & 79.92\%	& 80.34\% & 81.06\% & 90.23\% \\
\hline
50000 & 80.12\%	& 81.16\% & 82.12\%	& 91.56\% \\
\hline
\end{tabular}
\caption{Average accuracies of DAP and P-DAP}
\end{table}

Fig.7 gives the simulation experiment results of P-DAP. Fig.7 (a) shows both the regrets of P-DAP and DAP are sublinear over time. To be specific, we can see from the tendency of regret lines that as privacy preservation level get increased (smaller $\varepsilon $), regrets converged more slowly.  Fig.7 (b) shows
 our differentially private P-DAP has low-regret (no more than 0.03 per time slot) even for
a high level of privacy preservation (e.g., $\varepsilon $ = 0.01). The regret
obtained by the non-private algorithm has the lowest regret
as expected. More significantly, the regret gets closer to the
non-private regret as its privacy preservation is weaker.

Table III records our tested average accuracies for DAP and P-DAP with different privacy preservation level. As we can read from the table, average accuracy of DAP can reach to 91.56${\rm{\% }}$ and those of our P-DAP with different values of $\varepsilon $ is greater than 80${\rm{\% }}$ by time.

Fig. 8 shows our simulation results of GP-DAP and P-DAP, where we set $\varepsilon {\rm{ = }}0.01$. From Fig. 8 (a) tells us the regret of GP-DAP is less than that of P-DAP by $32{\rm{\% }}$. We can immediately draw the conclusion that GP-DAP cut the regret loss extensively. We also use different set of data with different volume to test the accuracies of P-DAP and GP-DAP. Fig. 8 (b) shows the comparison of these average accuracies. Both GP-DAP and P-DAP have high accuracy for each group, and the accuracies become slightly higher when increasing the sizes of groups. Obviously, GP-DAP always has higher accuracy than P-DAP.

 Table IV records the test result of GP-DAP and P-DAP ($\varepsilon {\rm{ = }}0.01$) of different user groups. At first glance, the accuracies increased slightly as we add more context samples into test group. This is due to the fact that, more samples can help systems get better estimation of each processing functions. Also, we can see that, the average accuracy of the GP-DAP will be greater than 88${\rm{\% }}$ as number of user arrivals exceed 30000.
 \begin{figure}
\begin{minipage}{0.5\linewidth}
\centering
\includegraphics[width=1.9in]{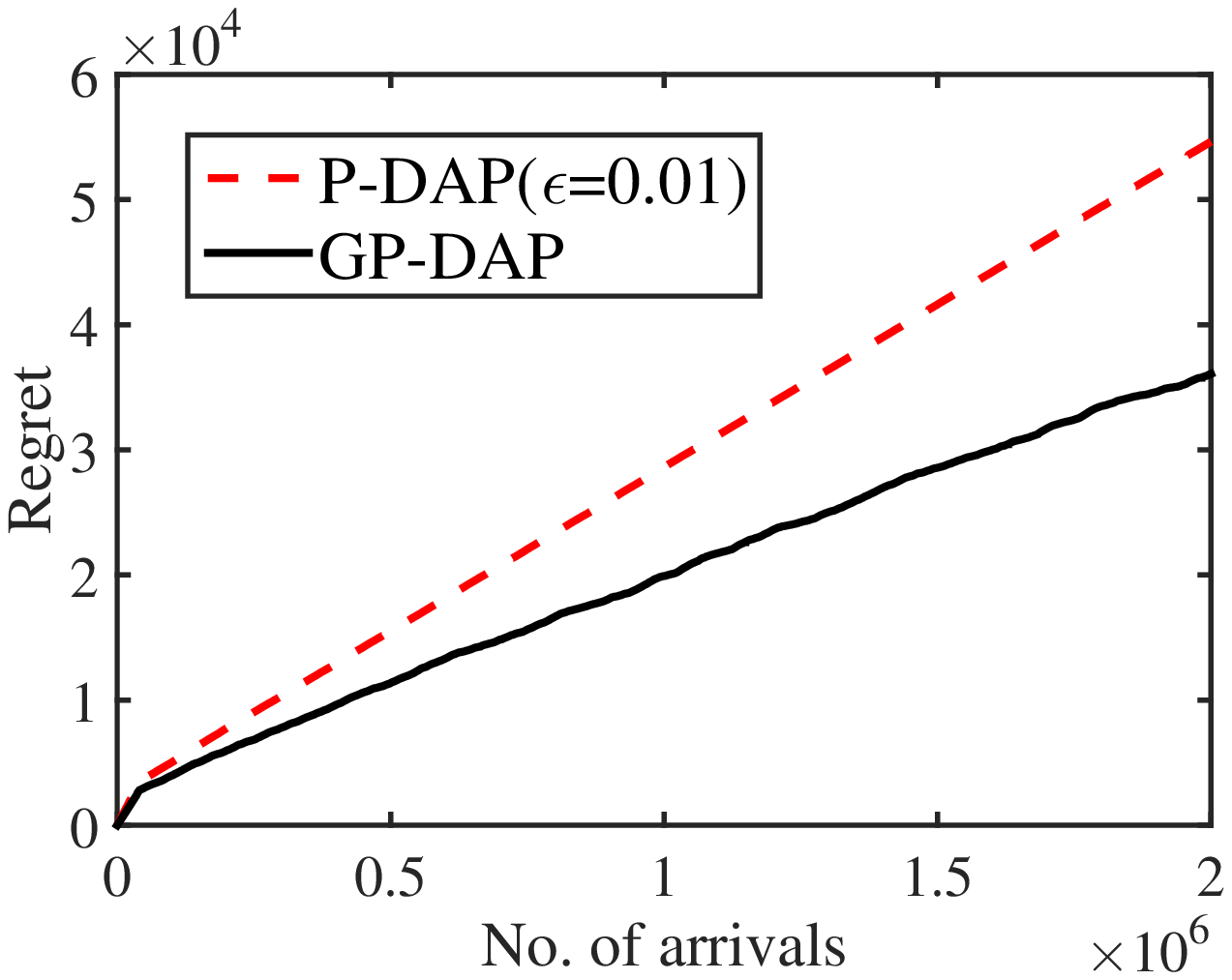}
\centering{(a) Regret}
\label{fig:side:a}
\end{minipage}%
\begin{minipage}{0.5\linewidth}
\centering
\includegraphics[width=1.9in]{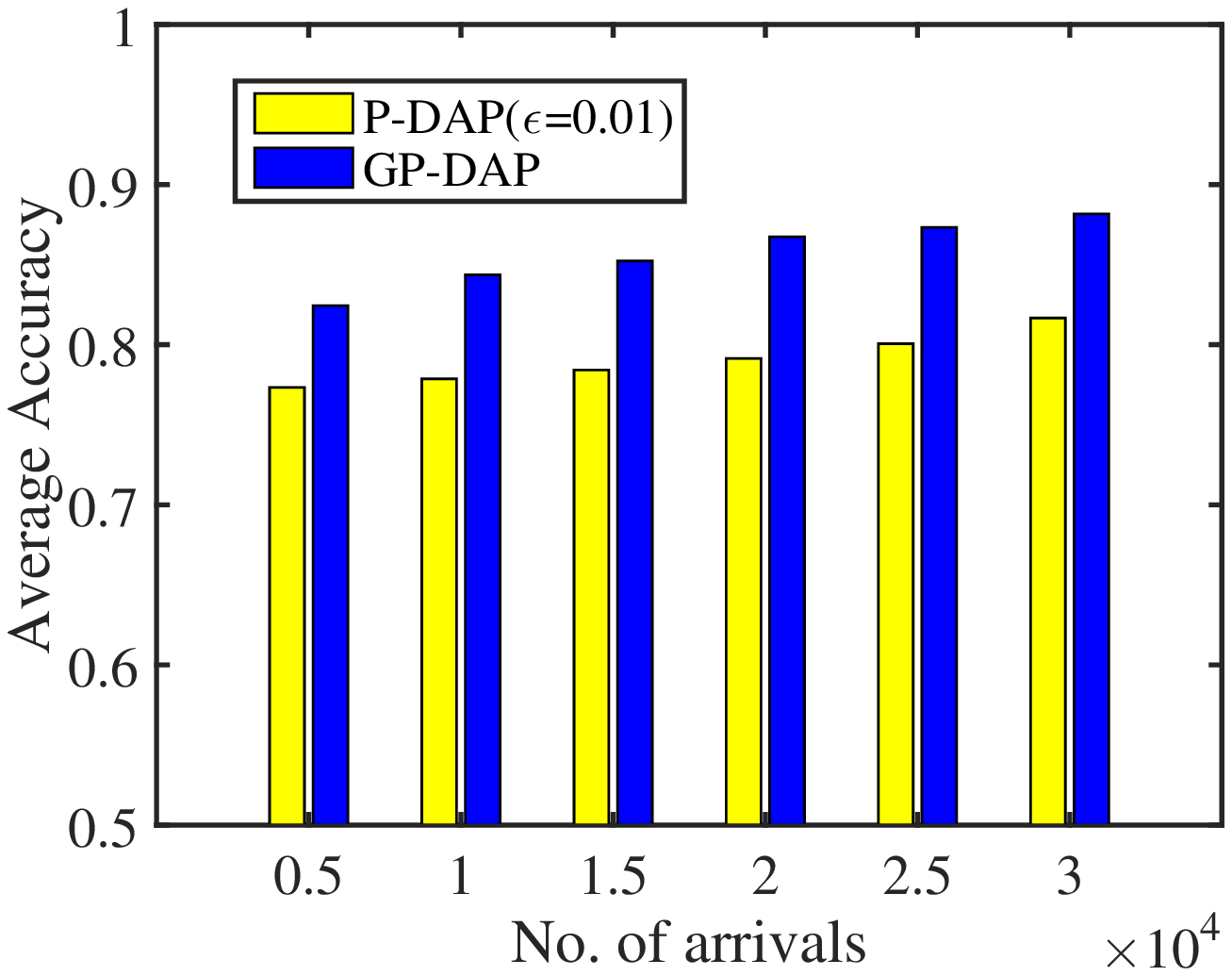}
\centering{(b) Average Accuracy}
\label{fig:side:b}
\end{minipage}
\caption{Regrets and Accuracies in P-DAP and GP-DAP}
\end{figure}

\begin{table}
 \vspace{-.5em}
\centering
\begin{tabular}{|c|c|c|c|c|c|c|}
\hline
\scriptsize N & \scriptsize 5000 & \scriptsize 10000 & \scriptsize 15000 & \scriptsize 20000 & \scriptsize 25000 & \scriptsize 30000 \\
\hline
\scriptsize GP-DAP & \scriptsize 82.44\% & \scriptsize 84.36\% & \scriptsize 85.23\% & \scriptsize 86.74\% & \scriptsize 87.33\% & \scriptsize 88.17\% \\
\hline
\multirow{2}{0.8cm}{\scriptsize P-DAP\\($\epsilon$=0.01)} & \multirow{2}{0.8cm}{\scriptsize 77.31\%} & \multirow{2}{0.8cm}{\scriptsize 77.87\%} & \multirow{2}{0.8cm}{\scriptsize 78.42\%}
& \multirow{2}{0.8cm}{\scriptsize 79.14\%} & \multirow{2}{0.8cm}{\scriptsize 80.06\%} & \multirow{2}{0.8cm}{\scriptsize 81.64\%} \\
&&&&&&\\
\hline
\end{tabular}
\caption{Tested average accuracies of GP-DAP and DAP}
\end{table}

\section{Conclusion}
In this paper, we have presented a differential private distributed learning framework for video recommendation for online social networks. To tackle with the large value and heterogeneity of big data, we adopt dynamic space partition to distributed contextual bandit. Concerned with the privacy of social network users and that of video service vendors, we use exponential mechanism and Laplace mechanism simultaneously. Furthermore, to alleviate the performance loss due to introducing differential privacy, we refine our framework to novel \emph{geometric differentially private} model. We have theoretically analyzed our algorithms in terms of performance loss (regret) and privacy preserving. We have also evaluated our algorithms, demonstrating their sublinear converged regrets, delicate trade-off between performance loss and privacy preserving level and extensively reduction.


%

\section*{Acknowledgment}
This research is supported by National Science Foundation of
China with Grant 61401169.

%
%

\ifCLASSOPTIONcaptionsoff
  \newpage
\fi



%

%

\begin{IEEEbiography}[{\includegraphics[width=1in,height=1.25in,clip,keepaspectratio]{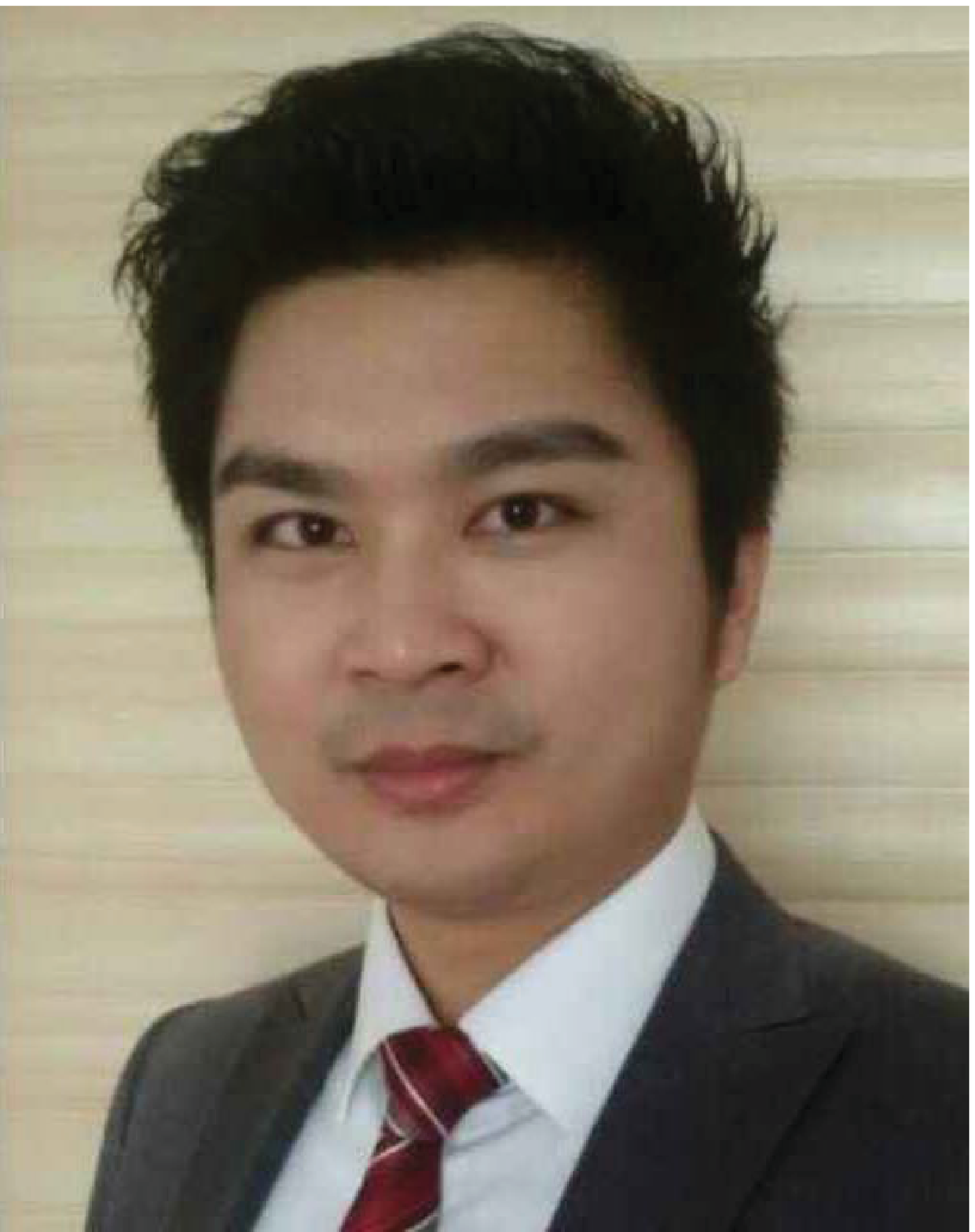}}]{Pan
Zhou (S'07--M'14)} is currently an associate professor
 with School of Electronic Information and Communications, Huazhong
University of Science and Technology, Wuhan, P.R. China. He received
his Ph.D. in the School of Electrical and Computer Engineering at the
Georgia
Institute of Technology (Georgia Tech) in 2011, Atlanta, USA. He
received his B.S. degree in the \emph{Advanced Class} of
HUST, and a M.S. degree in the Department of Electronics and
Information Engineering
from HUST, Wuhan, China, in 2006 and 2008, respectively.
He held honorary degree in his bachelor and merit research award
of HUST in his master study. He was a
senior technical member at Oracle Inc, America during 2011 to 2013,
Boston, MA, USA,  and worked on hadoop and distributed storage system
for big data
analytics at Oralce cloud Platform.  His current research interest
includes:  communication and information networks, security and
privacy,  machine learning and big data.
\end{IEEEbiography}

\begin{IEEEbiography}[{\includegraphics[width=1in,height=1.25in,clip,keepaspectratio]{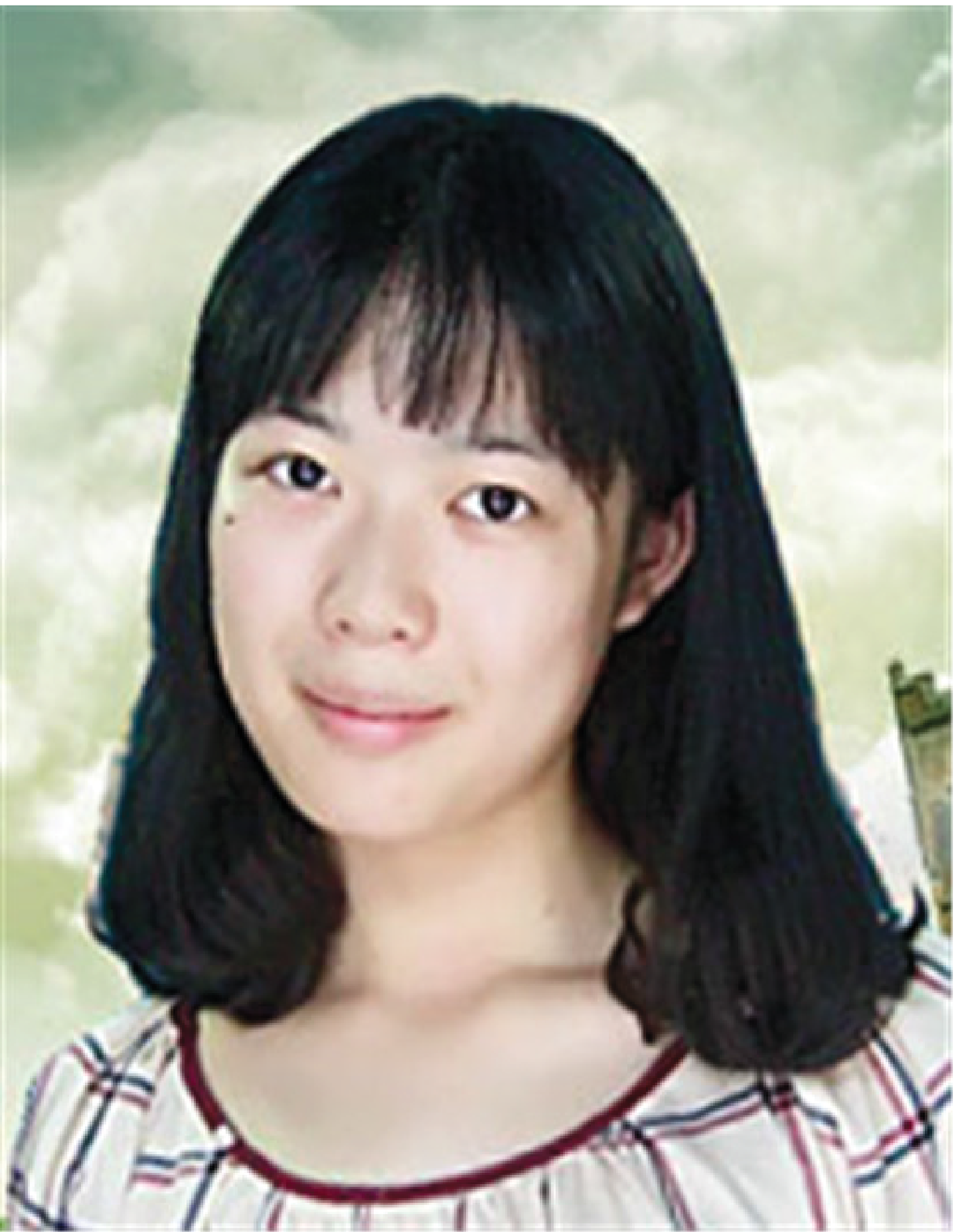}}]
{Yingxue Zhou} (S'15) is currently working toward the B.S. degree at the School
of Electronic Information and Communications,
Huazhong University of Science and
Technology, Wuhan, P.R. China. Her current research interests include:  online learning, Big Data analytics, differential privacy and social networks. She is a student member of the IEEE.

\end{IEEEbiography}

\begin{IEEEbiography}[{\includegraphics[width=1in,height=1.25in,clip,keepaspectratio]{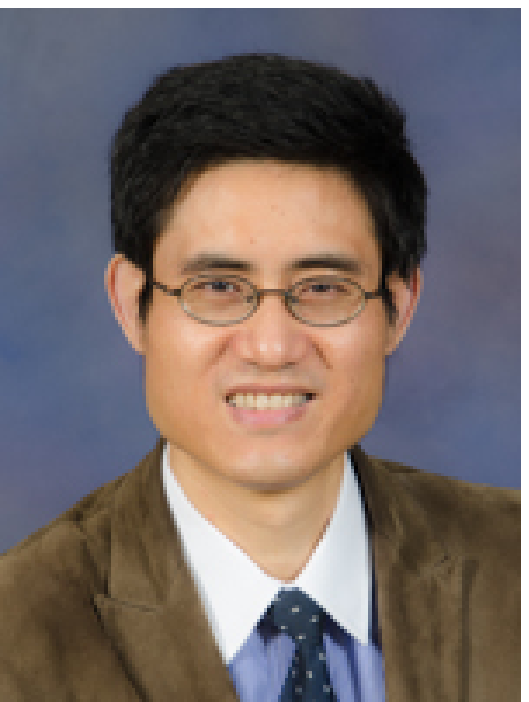}}]{Dapeng Wu}
(S'98-M'04-SM06-F'13) received
Ph.D. in Electrical and Computer Engineering from
Carnegie Mellon University, Pittsburgh, PA, in 2003.
He is a professor at the Department of Electrical
and Computer Engineering, University of Florida,
Gainesville, FL. His research interests are in the
areas of networking, communications, signal processing,
computer vision, machine learning, smart
grid, and information and network security.
\end{IEEEbiography}

\begin{IEEEbiography}[{\includegraphics[width=1in,height=1.25in,clip,keepaspectratio]{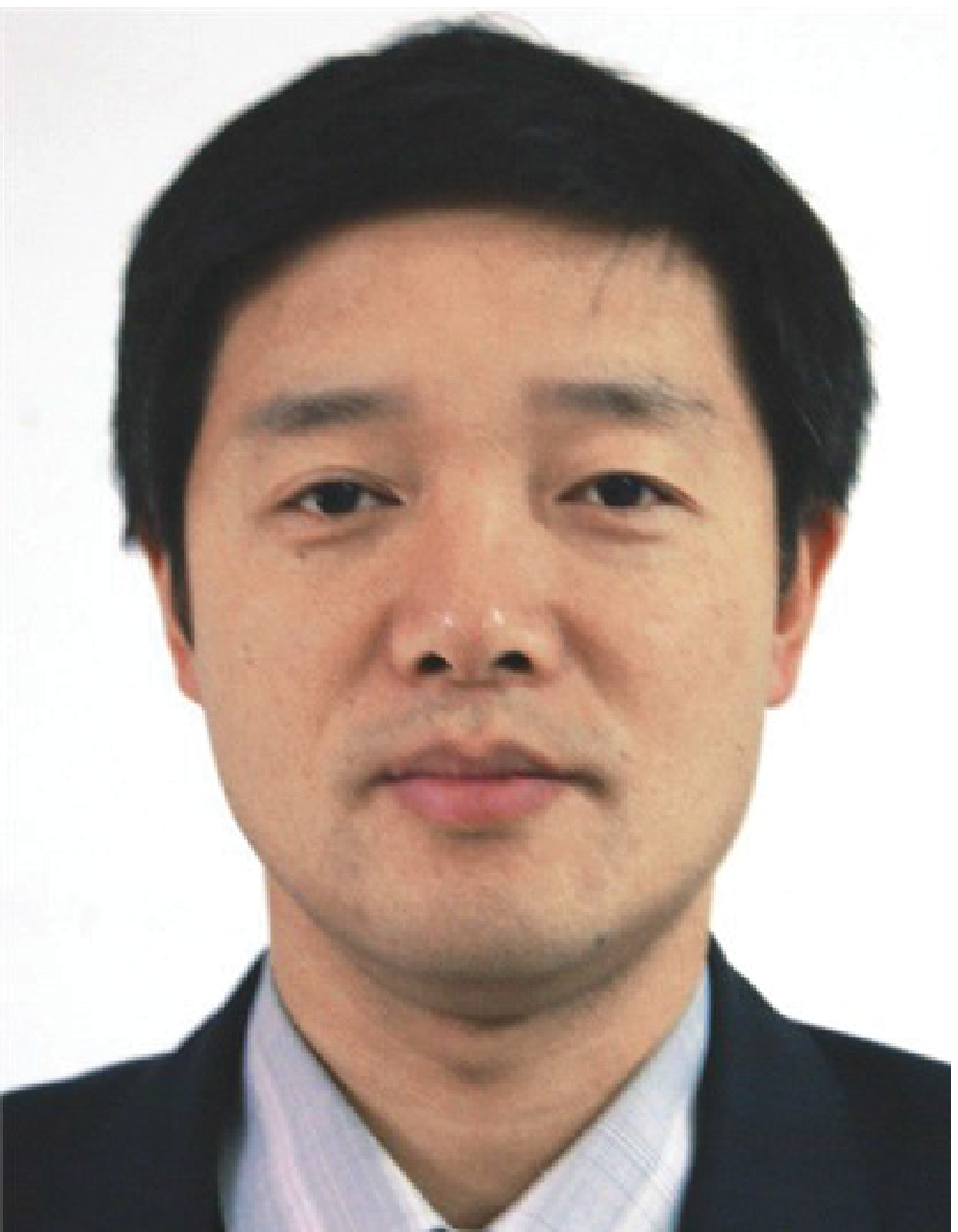}}]{Hai Jin}
 (M'99-SM'06) received the Ph.D. degree
in computer engineering from Huazhong University
of Science and Technology (HUST), Wuhan, China,
in 1994.
He was with The University of Hong Kong,
Hong Kong, between 1998 and 2000, and was
a Visiting Scholar at the University of Southern
California, Los Angeles, between 1999 and 2000. He
is a Professor of computer science and engineering
with HUST, where he is currently the Dean of the
School of Computer Science and Technology. He
is also with the Cluster and Grid Computing Laboratory and the Service
Computing Technology and System Laboratory of Ministry of Education,
HUST. He has coauthored 8 books and published over 300 research papers.
His research interests include computer architecture, virtualization technology,
cluster computing and grid computing, peer-to-peer computing, network
storage, and network security.
\end{IEEEbiography}






\end{document}